\documentclass[sigconf]{acmart}

\usepackage{multirow,makecell}
\usepackage{amsthm}
\usepackage{subcaption}
\usepackage{enumitem}
\usepackage{bbding}
\usepackage{balance}

\theoremstyle{definition}

\newcommand{\autocss}{{\textsc{AutoCL}}\xspace}

\usepackage{xcolor}
\usepackage{xspace}

\AtBeginDocument{%
  \providecommand\BibTeX{{%
    \normalfont B\kern-0.5em{\scshape i\kern-0.25em b}\kern-0.8em\TeX}}}

\copyrightyear{2024}
\acmYear{2024}
\setcopyright{acmlicensed}\acmConference[CIKM '24]{Proceedings of the 33rd ACM International Conference on Information and Knowledge Management}{October 21--25, 2024}{Boise, ID, USA}
\acmBooktitle{Proceedings of the 33rd ACM International Conference on Information and Knowledge Management (CIKM '24), October 21--25, 2024, Boise, ID, USA}
\acmDOI{10.1145/3627673.3680086}
\acmISBN{979-8-4007-0436-9/24/10}

\begin{document}

\title{Automated Contrastive Learning Strategy Search for Time Series}

\author{Baoyu Jing}
\orcid{0000-0003-1564-6499} 
\affiliation{
  \institution{University of Illinois}
  \city{Urbana-Champaign}
  \state{IL}
  \country{USA}
}
\email{baoyuj2@illinois.edu}

\author{Yansen Wang}
\orcid{0009-0005-5280-2050} 
\affiliation{
  \institution{Microsoft Research Asia}
  \city{Shanghai}
  \country{China}
}
\email{yansenwang@microsoft.com}

\author{Guoxin Sui}
\orcid{0000-0003-2033-1774} 
\affiliation{
  \institution{Microsoft Research Asia}
  \city{Shanghai}
  \country{China}
}
\email{guoxin.sui@illinois.edu}

\author{Jing Hong}
\orcid{0009-0006-7823-4669} 
\affiliation{
  \institution{Ruijin Hospital}
  \city{Shanghai}
  \country{China}
}
\email{hj40785@rjh.com.cn}

\author{Jingrui He}
\orcid{0000-0002-6429-6272} 
\affiliation{
  \institution{University of Illinois}
  \city{Urbana-Champaign}
  \state{IL}
  \country{USA}
}
\email{jingrui@illinois.edu}

\author{Yuqing Yang}
\orcid{0000-0003-3518-5212} 
\affiliation{
  \institution{Microsoft Research Asia}
  \city{Shanghai}
  \country{China}
}
\email{yuqing.yang@microsoft.com}

\author{Dongsheng Li}
\orcid{0000-0003-3103-8442} 
\affiliation{
  \institution{Microsoft Research Asia}
  \city{Shanghai}
  \country{China}
}
\email{dongsli@microsoft.com}

\author{Kan Ren}
\orcid{0000-0002-4032-9615} 
\affiliation{
  \institution{ShanghaiTech University}
  \city{Shanghai}
  \country{China}
}
\email{renkan@shanghaitech.edu.cn}

\authornote{Correspondence to Kan Ren.}
\renewcommand{\shortauthors}{Baoyu Jing et al.}

\begin{abstract}
In recent years, Contrastive Learning (CL) has become a predominant representation learning paradigm for time series.
Most existing methods manually build specific CL Strategies (CLS) by human heuristics for certain datasets and tasks.
However, manually developing CLS usually requires excessive prior knowledge about the data, 
and massive experiments to determine the detailed CL configurations.
In this paper, we present an Automated Machine Learning (AutoML) practice at Microsoft, which automatically learns CLS for time series datasets and tasks, namely Automated Contrastive Learning (\autocss). 
We first construct a principled search space of size over $3\times10^{12}$, covering data augmentation, embedding transformation, contrastive pair construction, and 
contrastive losses.
Further, we introduce an efficient reinforcement learning algorithm, which optimizes CLS from the performance on the validation tasks, to obtain effective CLS within the space.
Experimental results on various real-world datasets demonstrate that \autocss\ could automatically find the suitable CLS for the given dataset and task. 
From the candidate CLS found by \autocss\ on several public datasets/tasks, we compose a transferable Generally Good Strategy (GGS), which has a strong performance for other datasets.
We also provide empirical analysis as a guide for the future design of CLS.

\end{abstract}



\maketitle

\section{Introduction}
Time series have been collected and analyzed for plenty of real-world applications, e.g., clinical diagnosis \cite{eldeletime,chen2024eegformer}, electricity forecasting \cite{zhou2021informer,liu2024timesurl} and various monitoring tasks \cite{ren2019time,zhou2020data,jing2021network,jing2024casper,wang2023networked}.
A paramount challenge for time series representation learning lies in developing an efficacious encoder capable of deriving informative embeddings, 
which could be readily applicable to a variety of downstream tasks, e.g., classification \cite{zhang2022self,nie2022time,chen2024contiformer}, forecasting \cite{yue2022ts2vec,jing2022retrieval} and anomaly detection \cite{ren2019time,li2021outlier,zheng2024mulan}.

A predominant representation learning paradigm is Contrastive Learning (CL) \cite{liu2021self, zheng2024heterogeneous},
of which the core idea is to train models by pulling the positive embedding pairs closer and pushing the negative embedding pairs far apart \cite{jaiswal2020survey,chen2020simple,jing2021hdmi,zheng2022contrastive,feng2022adversarial,zheng2021deeper}.
The preponderance of existing research in CL as applied to time series primarily emphasizes the development of sophisticated data augmentation techniques and the construction of contrastive pairs \cite{liu2023self}. 
For instance, TS-TCC~\cite{eldeletime} employs jittering and scaling to produce varied perspectives of the input time series.
These generated perspectives, originating from the same data sample, are then treated as positive instances within the model's training regime.
TS2Vec~\cite{yue2022ts2vec} uses random cropping to generate positive pairs and treat other data points in the context as negative samples.
TF-C~\cite{zhang2022self} augments the input via time-domain and frequency-domain masking, and embeddings from different samples are treated as negative pairs.

Despite the effectiveness of these approaches, the majority are tailored with manual construction of Contrastive Learning Strategies (CLS) specific to particular datasets and tasks, representing distinct examples within the broader spectrum of general CLS space, which could contain millions even trillions of choices.
Crafting CLS for specific tasks or datasets in such a huge space through heuristic methods could be challenging, often demanding extensive domain expertise and considerable manual effort through numerous trial-and-error iterations.
A natural question arises: is it possible to build a system to automatically find a suitable CLS for a given task?
A pertinent question emerges: can we identify a CLS that is generally effective across a dynamic range of tasks and datasets?

Aiming at automating the process of machine learning, Automated Machine Learning (AutoML) \cite{he2021automl} has become popular in recent years.
In the field of CL, some works optimize data augmentation strategies on images \cite{tian2020makes} and graphs \cite{you2021graph,feng2024ariel,yin2022autogcl}.
For time series, a recent work InfoTS \cite{luo2023time} searches for optimal data augmentations based on information theory.
These studies predominantly focus on data augmentation, but often overlook other crucial dimensions of CL, such as loss functions, the construction of contrastive pairs, and embedding transformations, which are typically crafted by human intervention.
Specifically, a CLS encompasses diverse loss functions (e.g., InfoNCE \cite{oord2018representation} and Triplet Loss \cite{hoffer2015deep}) and metric functions (e.g., dot product, cosine similarity, negative Euclidean distance) that affect representation learning quality \cite{zheng2022contrastive,liu2021self,jing2022coin,yan2024reconciling,jing2024sterling}. 
Contrastive pair construction \cite{oord2018representation,yue2022ts2vec,eldeletime,jing2022x,zheng2023fairness,li2022graph} also plays a key role in distinguishing sample-level characteristics or temporal dependencies.
Embedding techniques, including augmentation \cite{wang2019implicit} and normalization \cite{wang2020understanding}, further influence the pretrained model performance.
Few works have been conducted on jointly considering all these aspects of the contrastive learning on time series.

To reduce the burden of excessive prior domain knowledge and massive manual trials of CL on time series for specific tasks, in this paper, we present an \underline{Auto}mated \underline{C}ontrastive \underline{L}earning (\autocss) framework at Microsoft, which aims at learning to contrastively learn the representations for various time series tasks.
Specifically, we first construct a comprehensive solution space of CLS, which not only covers the most important configurable dimensions of CL,
but also considers the modest range of options for each dimension.
The whole space is comprised of dimensions including data augmentations, embedding transformations, contrastive pair construction, and contrastive learning objectives, with up to $10^{12}$ level of options in total.
To efficiently target the suitable CLS in such a huge solution space, we further introduce a reinforcement learning algorithm which directly optimizes the CL performance on downstream tasks in validation datasets.
Compared with the recent automation-based CL method InfoTS \cite{luo2023time}, our method takes the two-fold advantage from more comprehensive solution space and the direct optimization approach on CL, while InfoTS only searches data augmentation configurations based on task-agnostic criteria, which does not consider the intrinsic connection between CLS and the downstream task, thus may result in targeting an improper CLS.

We empirically evaluate \autocss and the derived CLS over three downstream tasks including classification, forecasting, and anomaly detection on real-world time series datasets.
We first directly apply \autocss over each task to obtain its suitable CLS, which could significantly outperform existing CL methods, demonstrating the superiority of \autocss.
Then we derive a Generally Good Strategy (GGS) based on the candidate CLS found by \autocss from several tasks and test GGS on all tasks.
The results show that GGS could achieve remarkable performance for all tasks, showing its strong transferability across tasks and datasets. 
We also provide extensive empirical analysis on the relationship between the candidate CLS and the model's performance on downstream tasks, hoping to provide guidance for the future design of CLS.
We also tested our solution in a real-world application of epilepsy seizure detection task in Shanghai Ruijin Hospital, which has generally good performance. 
Our major contributions are summarized as follows:
\begin{itemize}[leftmargin=3mm]
    \item We introduce \autocss to automatically derive suitable CLS for various tasks and datasets on time series.
    A comprehensive solution space covering the most critical dimensions of CL with an efficient reinforcement learning algorithm has been proposed.
    \item Extensive experiments in both public benchmarks and a deployed application have demonstrated the superiority of the derived CLS with better performance compared to the existing CL methods. 
    \item From the candidate CLS, we first provide some empirical findings on CLS and downstream tasks to guide the future design CLS. We also obtain a Generally Good Strategy (GGS), which can be used as a strong baseline for various time series datasets and tasks.
\end{itemize}
\section{Methodology}
In this section, we first formulate the problem of the automated contrastive learning strategy search and then introduce a method called \autocss for the problem. We will elaborate the two components of \autocss: the solution space and the search algorithm.

\begin{figure}
    \centering
    \includegraphics[width=.3\textwidth]{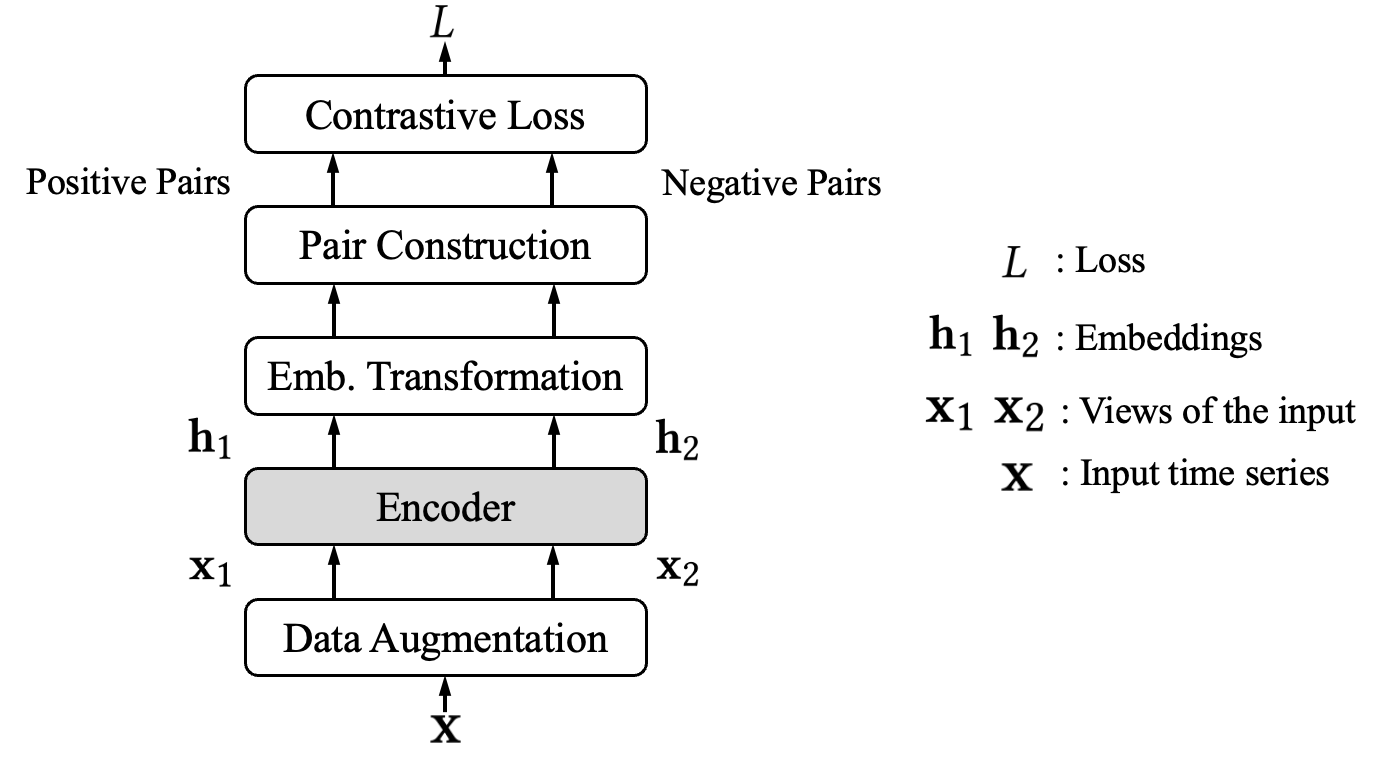}
    \caption{Illustration of Contrastive Learning Strategy (CLS).}
    \label{fig:space}
    \vspace{-5pt}
\end{figure}

\subsection{Problem Formulation}
Denote $\mathcal{D}$ as a time series dataset comprised of training $\mathcal{D}_\text{train}$, validation $\mathcal{D}_\text{val}$ and test $\mathcal{D}_\text{test}$ sets;
$\mathcal{T}$ as a downstream task along with evaluation metrics $\mathcal{M}$;
$f_{E}$ as the time series encoder.
The problem of automated Contrastive Learning Strategy (CLS) search has two sub-problems:
(1) define a solution space $\mathcal{S}$;
(2) build a search algorithm $f$ to find a suitable CLS $A\in\mathcal{S}$ based on the validation performance on $\mathcal{D}_\text{val}$ regarding $\mathcal{M}$, s.t., after pre-training with $A$ on $\mathcal{D}_\text{train}$, $f_E$ will have a good performance on $\mathcal{D}_\text{test}$.

\subsection{Solution Space}\label{sec:space}
Based on the literature and our own experiences, the space is defined by following the principles \cite{you2020design}: 
(1) encompasses the key dimensions of CLS based on prior human-designed strategies;
(2) accounts for a moderate spectrum of choices for each dimension.

An illustration of a CLS is shown in Figure \ref{fig:space}.
The essence of CL is to extract semantic-preserving embeddings invariant to small data perturbations \cite{tian2020makes}, and thus data augmentation is the cornerstone dimension of CLS.
The commonly used practices to enhance the robustness of representation are embedding transformations, e.g., embedding augmentation \cite{wang2019implicit} and normalization.
There are various aspects to construct positive and negative embedding pairs, where different aspects reflect different characteristics of data.
Different tasks might place emphasis on different aspects.
Finally, different similarity functions and different CL losses could have divergent impacts on the learned representations.
In summary, we consider four dimensions: data augmentations, embedding transformations, contrastive pair construction, and contrastive losses with around $3\times10^{12}$ options in total.
A summary of the dimensions, sub-dimensions, and their options is shown in Table \ref{tab:space}, and we describe the details in the following content.

\begin{table}[t]
    \centering
    \scriptsize
    \caption{The proposed solution space.}
    \begin{tabular}{c|c|c}
    \toprule
    Dimensions &  Sub-Dimensions & Options \\
    \midrule
    \multirowcell{7}{Data\\ Augmentations} & Resizing (Length) \cite{mehari2022self} & \multirow{6}{*}{$0.0, 0.1, \cdots, 0.9, 0.95$}\\
     & Rescaling (Amplitude) \cite{han2021semi} & \\
     & Jittering \cite{eldeletime}& \\
     & Point Masking \cite{zerveas2021transformer}& \\
     & Frequency Masking \cite{zhang2022self}& \\
     & Random Cropping \cite{yue2022ts2vec}& \\
     \cmidrule{2-3}
     & Order & 0, 1, 2, 3, 4\\
    \midrule
    \multirowcell{3}{Embedding\\ Transformations} & Jittering \cite{wang2019implicit} &  \multirow{2}{*}{$0.0, 0.1, \cdots, 0.9, 0.95$}\\
     & Masking & \\
    \cmidrule{2-3}
     & Normalization & None, LayerNorm \cite{ba2016layer}, $l_2$\\
    \midrule
    \multirowcell{6}{Contrastive Pair\\Construction} & Instance Contrast \cite{eldeletime} & True\\
     & Temporal Contrast \cite{yue2022ts2vec} & True, False\\
     & Cross-Scale Contrast \cite{bachman2019learning} & True, False\\
     \cmidrule{2-3}
     & Kernel Size & 0, 2, 3, 5\\
     & Pooling Operator & Avg, Max\\
     \cmidrule{2-3}
     & Use Adjacent Neighbor \cite{franceschi2019unsupervised} & True, False\\
    \midrule
    \multirowcell{3}{Contrastive\\ Losses} & Loss Types & InfoNCE \cite{oord2018representation}, Triplet \cite{hoffer2015deep}\\
     & Similarity Functions & Dot, Cosine, Euclidean Distance\\
     & Temperature Parameters & $10^{-2}, 10^{-1}, 10^{0}, 10^{1}, 10^{2}$\\
    \bottomrule
    \end{tabular}
    \label{tab:space}
    \vspace{-5pt}
\end{table}

\textbf{{Data Augmentations.}}
Data augmentations transform the input data into different but related views, which are the cornerstones of a CLS.
We consider 6 most commonly adopted data augmentations as the sub-dimensions, including \textbf{resizing (length)} \cite{mehari2022self}, \textbf{rescaling (amplitude)} \cite{han2021semi}, \textbf{jittering} \cite{eldeletime}, \textbf{point masking} \cite{zerveas2021transformer}, \textbf{frequency masking} \cite{zhang2022self}, and \textbf{random cropping} \cite{yue2022ts2vec}.
Each augmentation is associated with a parameter $p\in[0,1]$.
Let $\mathbf{x}\in\mathbb{R}^{T\times c}$ be an input time series, where $T$ and $c$ are the length and the number of variables.
For example, $p$ in the point masking refers to the ratio of the input data points to be masked: $\mathbf{x}=\mathbf{x}\odot\mathbf{m}$, where $\mathbf{m}[t]\sim Bernoulli(1-p)$.
For more details of each augmentation, please refer to the corresponding paper.
We discretize the value of $p$ into 11 choices: $\{0.0, 0.1, \cdots, 0.8, 0.9, 0.95\}$.
For the first value, we use $0.0$ to represent the setting where a specific data augmentation is not enabled.
For the last value, we do not use $1.0$ since it is meaningless for some data augmentations in practice.
For example, in {point masking}, $p=1.0$ means masking out all the input data.

Additionally, the \textbf{order} of applying data augmentations also 
influences the learned embeddings \cite{cubuk2019autoaugment}.
For instance, the outcome of applying {point masking} followed by {random cropping} differs from first applying random cropping and then point masking.
Therefore, we design 5 different orders of applying data augmentations.
In summary, there are totally $5\times11^6$ options for data augmentations.

\textbf{{Embedding Transformations.}}
Embedding augmentations \cite{wang2019implicit} and normalization \cite{wang2020understanding} have been explored and proven useful in the literature.
For embedding augmentations, we consider \textbf{embedding jittering} \cite{wang2019implicit} and \textbf{embedding masking} as the sub-dimensions.
Similar to the {jittering} and {point masking} in data augmentations, they are also associated with a parameter $p\in[0,1]$, whose value is also discretized into 11 values.
For \textbf{normalization}, we consider to use no norms, $l_2$ norm, and layer norm \cite{ba2016layer} as the options. 
In summary, there are $3\times11^2$ options for embedding transformations.

\textbf{{Contrastive Pair Construction}.}
There are two types of embedding pairs: positive pairs, where embeddings are semantically related to each other, and negative pairs, where embeddings do not necessarily share hidden semantics.
For time series, contrastive pairs can be constructed from three aspects: \textbf{instance contrast} \cite{eldeletime}, \textbf{temporal contrast} \cite{yue2022ts2vec}, and \textbf{cross-scale contrast} \cite{bachman2019learning}.

Let $\mathbf{h}_1,\mathbf{h}_2\in\mathbb{R}^{T\times d}$ be the embeddings of two views $\mathbf{x}_1$, $\mathbf{x}_2$ of an input $\mathbf{x}$, where $T$ and $d$ are the length and dimension size.
The {instance contrast} aims to distinguish embeddings of two different time series instances $\mathbf{x},\mathbf{x}'$ by pulling $\mathbf{h}_1$ and $\mathbf{h}_2$ (also $\mathbf{h}_1'$ and $\mathbf{h}_2'$) closer and pushing $\mathbf{h}_1,\mathbf{h}_2$ far away from $\mathbf{h}_1',\mathbf{h}_2'$.
The {temporal contrast} aims to preserve the local temporal information. 
For a time step $t$, it pulls $\mathbf{h}_1[t]$ and $\mathbf{h}_2[t]$ closer and \emph{optionally} also $\mathbf{h}_1[t]$ and its \textbf{temporally adjacent embeddings} $\mathbf{h}_1[t-1],\mathbf{h}_1[t+1]$ closer.
It pushes $\mathbf{h}_1[t]$ away from $\mathbf{h}_1[t']$, where $t'\notin\{t-1,t+1\}$. 
These two contrasts might be sufficient for short time series, e.g., $T<100$, however, they are incapable of capturing multi-scale dynamics of long time series, e.g., $T>1000$ \cite{yue2022ts2vec}.
We follow \cite{yue2022ts2vec} and apply \textbf{hierarchical pooling} over $\mathbf{h}_1,\mathbf{h}_2$ to obtain multi-scale embeddings $\mathbf{h}_1^{(s)},\mathbf{h}_2^{(s)}\in\mathbb
{R}^{T_s\times d}$, where $s\in\{1,\cdots,S\}$ is the scale index and $T_s$ is the length.
The hierarchical pooling is associated with two parameters: \textbf{pooling operator} (average or max pooling) and \textbf{kernel size} ($\{0,2,3,5\}$).
The above instance contrast and temporal contrast are applied for all the scales.
In addition, \textbf{cross-scale contrast} \cite{bachman2019learning} could further improve the cross-scale consistency.
Specifically, if $\mathbf{h}^{(s+1)}[t_{s+1}]$ is pooled from $\mathbf{h}^{(s)}[t_s]$, then they should be pulled closer, otherwise, should be pushed away from each other.

In practice, one must choose at least one contrastive aspect for CL.
Based on prior studies and our experiences, the instance contrast is the most indispensable aspect, which is always included.
As a result, there are 2=3-1 contrast aspects, 2 pooling operators, 4 kernel sizes and 2 options (include temporal adjacent embeddings or not) for the temporal contrast, which means we have 32 options in total.

\textbf{{Contrastive Losses.}}
Different \textbf{types} of contrastive losses behave differently.
For example, InfoNCE \cite{oord2018representation} guides the model's learning by maximizing the probabilities of the positive pairs, and Triplet loss \cite{hoffer2015deep} maximizes the similarity score of the positive pairs.
A contrastive loss is usually comprised of a \textbf{similarity function}, measuring the similarity of embeddings, and a \textbf{temperature parameter} \cite{oord2018representation}, controlling the magnitude of similarity scores.
For the similarity function, we consider dot product, cosine, and negative Euclidean distance.
For the temperature parameter, 
we consider $\{10^{-2},10^{-1},10^{0},10^{1},10^{2}\}$.
In summary, there are 2 loss types, 3 similarity functions, and 5 temperatures, resulting in 30 options.

\subsection{Search Algorithm}
As shown in Section \ref{sec:space}, there are around $3\times10^{12}$ total strategies in the solution space.
Therefore, it could be extremely challenging and resource-consuming to manually pick a suitable CLS from the entire space.
In this paper, we aim to automate the process of searching the suitable CLS within the solution space.
In the following content, we first present an overview of the algorithmic part of \autocss, and then elaborate its details, including the controller network and the two phases: candidate search and candidate evaluation.

\begin{figure}
    \centering
    \includegraphics[width=.3\textwidth]{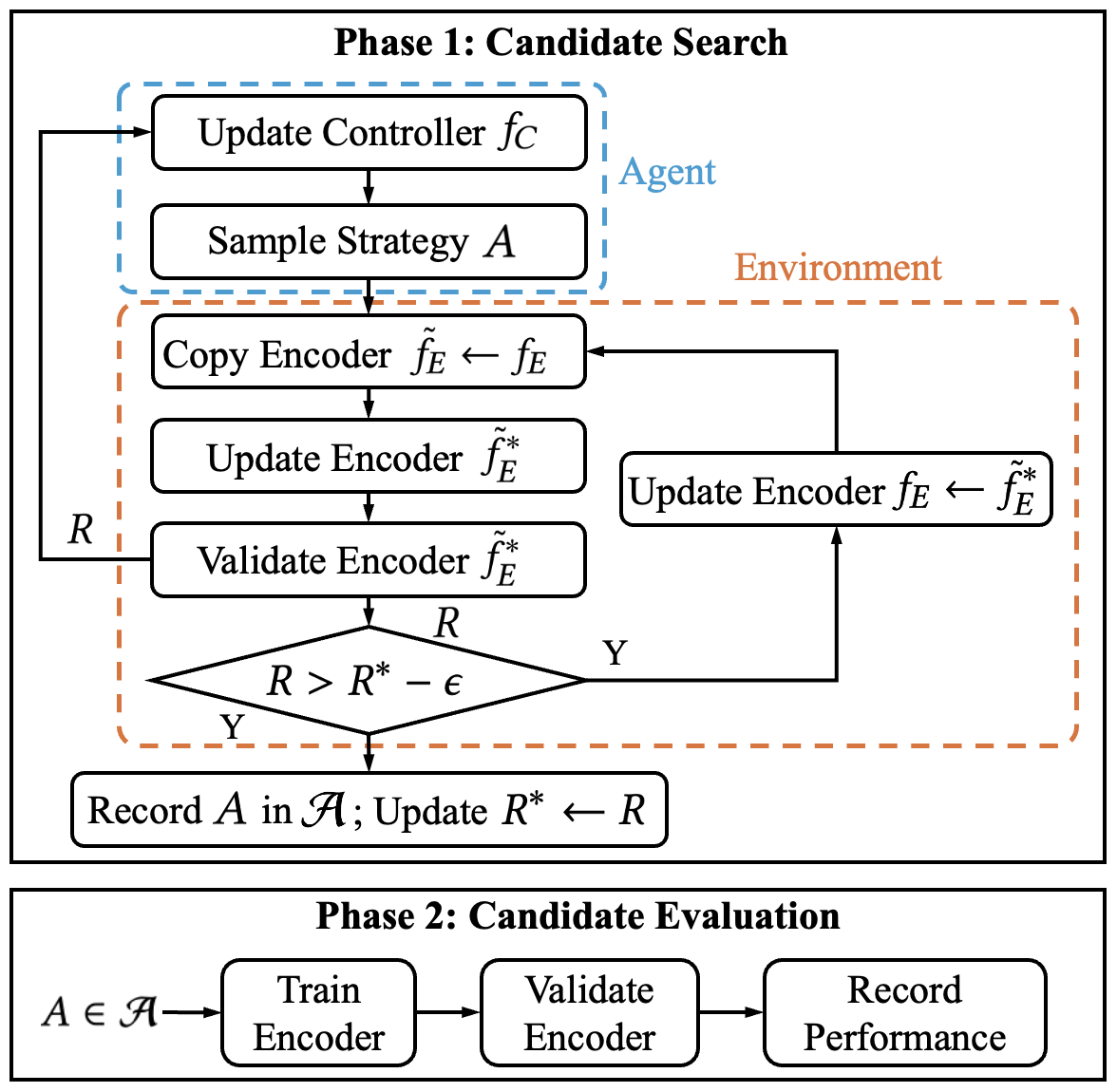}
    \caption{Illustration of the search algorithm of \autocss. 
    }
    \label{fig:overview}
    \vspace{-5pt}
\end{figure}

\textbf{{Overview.}}
In recent years, Reinforcement Learning (RL) has become a popular paradigm to implement AutoML pipelines.
Our proposed algorithm follows the RL paradigms in \cite{cubuk2019autoaugment,li2022autolossgen,zhao2021automated}, and an illustration is shown in Figure~\ref{fig:overview}.
The search algorithm has two phases: \emph{candidate search} and \emph{candidate evaluation}.
During candidate search, the agent, i.e., controller network $f_C$, interacts with the environment, including encoder training and validation, etc., for several iterations.
For each iteration, $f_C$ samples a CLS $A$ and sends it to the environment.
Then, the environment trains the encoder $f_E$ based on $A$ and returns a reward $R$ to $f_C$.
Finally, $f_C$ is updated based on $R$, and $A$ satisfying certain requirements is recorded in the candidate set $\mathcal{A}$.
During candidate evaluation, for each candidate $A\in\mathcal{A}$, we train $f_E$ based on $A$, then evaluate $f_E$ on the downstream task and record its performance on the validation data.

\begin{figure}[t!]
    \centering
    \includegraphics[width=.3\textwidth]{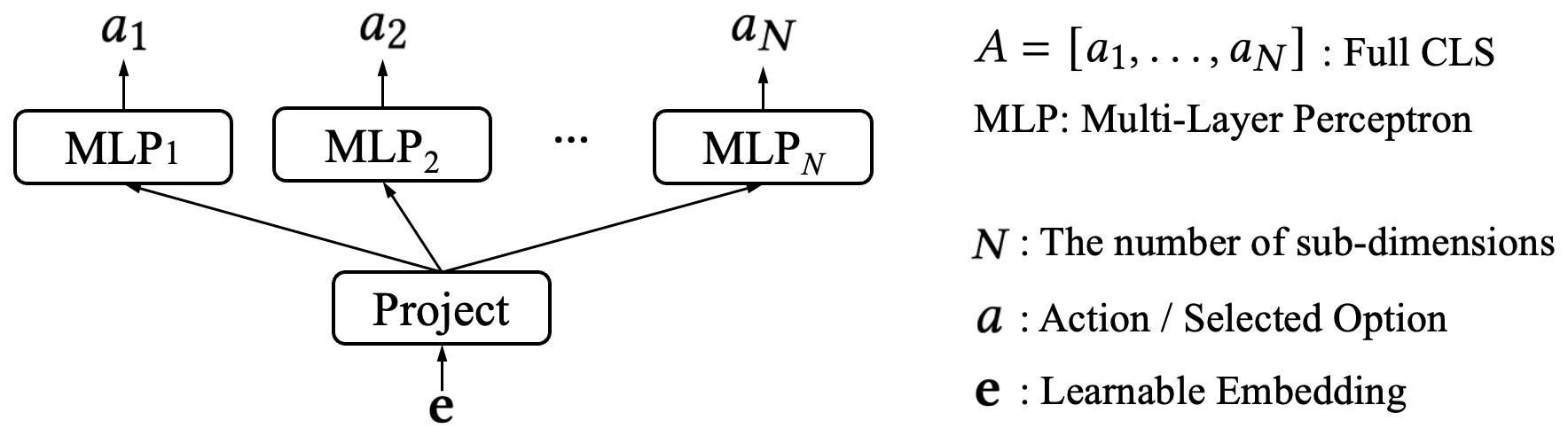}
    \caption{Controller network $f_C$.}
    \label{figs:controller}
    \vspace{-5pt}
\end{figure}

\textbf{{Controller Network}.}
The controller network $f_C$ learns to sample CLS for the given dataset and downstream task, 
an illustration of which is shown in Figure \ref{figs:controller}.
$f_C$ is comprised of 
(1) a learnable embedding $\mathbf{e}\in\mathbb{R}^{d}$, which learns the information of the given dataset and task; 
(2) a project function, projecting $\mathbf{e}$ into a hidden space; 
(3) $N$ action branches, each of which samples an option, e.g., 0.5, for its corresponding sub-dimension, e.g., point masking.
Formally, $a_n$ is sampled according to the probability $\mathbf{p}_n = \text{MLP}_n(\sigma(\mathbf{W}\mathbf{e}))\in\mathbb{R}^{n_K}$,
where $\mathbf{W}\in\mathbb{R}^{d\times d}$ and $\sigma$ are the weight and the tanh activation of the project layer.
We denote a full strategy as $A=\{a_1, a_2, \dots, a_N\}$.

\textbf{{Candidate Search.}}\label{sec:candidate_search}
During candidate search, parameters of both the controller $f_C$ and the encoder $f_E$ need to be updated.
Optimizing their parameters $\theta_C$ and $\theta_E$ is a bi-level optimization problem  \cite{zhao2021automated}:
\begin{equation}\label{eq:bi_level}
\centering
\begin{split}
    \max_{\theta_C}\, &\mathbb{E}_{A\sim f_C}[R(A, \theta_E^*)]\\
    \text{s.t.} \,&\theta_{E}^*(A) = \arg\min_{\theta_E} L(\theta_E, A).
\end{split}
\end{equation}
Given the strategy sampled by the controller $A\sim f_C$, 
the lower-level problem is to find the optimal $\theta_E^*$ that minimizes the contrastive loss $L\in A$.
The upper-level problem is to maximize the reward $R$ on the validation set by optimizing $\theta_C$.
In practice, optimizing $\theta_C$ after the completion of optimizing $\theta_E$, i.e., conducting the complete CL for $f_E$, can be extremely costly \cite{zhao2021automated, li2022autolossgen}.
Following \cite{liu2018darts}, we use the first-order approximation of the gradient for $f_E$.
Let $\eta$ be the learning rate in gradient decent, then $\theta_E^*(A)$ is approximated by:
\begin{equation}\label{eq:first_order_approx}
    \theta_E^*(A)\approx \theta_E - \eta\bigtriangledown_{\theta_E}L(\theta_E,A) ~.
\end{equation}
After obtaining $\theta_E^*(A)$ (or $f_E^*$), the next step is to obtain the reward $R(A, \theta_E^*)$.
We first combine $f_E^*$ with a downstream model $f_D$ to obtain the full model $f=f_D\circ f_E^*$.
Next, we use the training data to train $f$ on the downstream task (e.g., classification), and we use the performance (e.g., accuracy) of $f$ on the validation data as the reward $R(A, \theta_E^*)$.
However, the raw reward $R$ cannot be directly used to update the parameters of the controller $\theta_C$, since for many metrics (e.g., accuracy or mean squared error), $R>0$ always holds true.
Without a negative reward, the agent might be unable to learn meaningful behaviors  \cite{williams1992simple}.
To address this problem, we use the maximum reward of previous $(n-1)$ steps $R^*=\max\{R_1,\dots,R_{n-1}\}$ as the baseline for the $n$-th step.
Therefore, the final reward $\Delta_{n}$ is:
\begin{equation}\label{eq:final_reward}
    \Delta_{n}=\alpha\cdot(R_{n}-R^* + \epsilon)
\end{equation}
where $\alpha > 0$ is a scaling factor and $\epsilon>0$ is a small constant.
Given $\Delta$, $\theta_C$ is updated based on the REINFORCE algorithm \cite{williams1992simple}.

In the bi-level optimization, if $\theta_E$ in the lower-level optimization is trained along an undesired direction, it will take many more steps to bring it back to the desired direction \cite{li2022autolossgen}.
To address this issue, rather than directly updating $\theta_E$, we make a copy of $\theta_E$, denoted by $\Tilde{\theta}_E$, and update the copied parameters to  $\Tilde{\theta}_E^*$.
Then we can obtain the raw $R(A, \Tilde{\theta}_E^*)$ and final $\Delta$ rewards based on $\Tilde{\theta}_E^*$.
If $\Delta<0$, then we discard $\Tilde{\theta}_E^*$ since it goes in the undesired direction.
If $\Delta>0$, then we replace the original encoder $\theta_E$ with $\Tilde{\theta}_E^*$, and also add $A$ to the candidate set $\mathcal{A}$, as $\theta_E$ is updated in the desired direction.

\textbf{{Candidate Evaluation.}}
To evaluate each candidate $A\in\mathcal{A}$, 
we first pre-train the encoder $f_E$ on the pre-training data based on $A$. 
Then we train a full model $f=f_D\circ f_E$ on the training data of the downstream task, e.g., classification, where $f_D$ is the downstream model.
The evaluation score, e.g., accuracy, of $f$ on the validation data is recorded as the performance of $A$.

\textbf{{Complexity Analysis.}}
\autocss is efficient in both phases.
In phase 1, we use the 1st-order approximation rather than exhausted training;
in phase 2, we can take advantage of parallel computing.
Let $N$ be the number of iterations in phase 1, $T_{t}$, be the time of training the encoder for one epoch, and $T_{v}$ be the time of validation.
The time complexity for phase 1 is $O(NT_{t}+(N+1)T_{v})$.
Let $M$ be the number of iterations of the full contrastive pre-training, $|\mathcal{A}|$ be the total number of candidates, and $K$ be the number of machines, then the time complexity of phase 2 is $O(\frac{|\mathcal{A}|}{K}(MT_{t}+T_{v}))$.
Therefore, the total complexity is $O((N+\frac{|\mathcal{A}|}{K}M)T_{t} + (N+1+\frac{|\mathcal{A}|}{K})T_{v})$.
In comparison, if we do not use the 1st order approximation in Equation \eqref{eq:first_order_approx}, then the complexity is $O(NMT_{t}+(N+1)T_{v})$ for phase 1, which is significantly larger than our current practice.
\section{Experiments}
\subsection{Experimental Setup}
\textbf{Datasets.} We use 7 public and 1 private datasets.
The \textit{public} datasets can be grouped by three tasks: classification, anomaly detection and forecasting.
For \textit{classification}: we use \textbf{HAR} \cite{anguita2013public}, \textbf{Epilepsy} \cite{andrzejak2001indications,eldeletime}.
HAR contains 10,299 sensor readings with a length of 128, where each reading corresponds to one of the 6 activities.
Epilepsy contains 11,500 electroencephalogram (EEG) time series segments with a length of 178 from 500 subjects, and the task is to recognize epileptic seizures.
For \textit{anomaly detection}, we use \textbf{Yahoo} \cite{laptev2015benchmark} and \textbf{KPI} \cite{ren2019time}.
{Yahoo} has 367 hourly time series with human-labeled anomaly points, which is web traffic data to Yahoo! services.
{KPI} contains 58 minutely KPI curves from various Internet companies.
These two datasets are originally split into 50\%/50\% for train/test.
During searching, we further split the training data by 90\%/10\% for training and validation.
For \textit{forecasting}, we use 
\textbf{ETTh1/h2/m1}/~\cite{zhou2021informer}, which are 2-year time series of oil temperature of electricity transformers.
The \textit{private} dataset contains \textbf{SEEG} (stereo-electroencephalogram) signals of 15 anonymized patients collected from Ruijin Hospital.
For each patient, 12-15 invasive electrodes each with 12-18 contacts are used to obtain about 3.5 hour readings.
Two neurosurgery technicians jointly marked the onset and end times of the seizures. 
The signal frequency is unified to 256Hz, and the time series are divided into segments length of 0.5s. 
The segments containing epileptic waves are considered as positive samples accounting for 1.6\% of the data.
Following \cite{andrzejak2001indications}, we balance the two classes into 50\%:50\%, resulting in 11,778 samples.
We split the data into 70\%/10\%/20\% for train/validation/test for each patient.


\textbf{Comparison Methods.}
(1) Automatic data augmentation learning method: InfoTS \cite{luo2023time}.
(2) Handcrafted CL methods: InfoTS \cite{luo2023time}, TS2Vec \cite{yue2022ts2vec}, TS-TCC \cite{eldeletime}, CoST \cite{woo2021cost}, TNC \cite{tonekaboni2020unsupervised}, CPC \cite{oord2018representation} and Self-EEG \cite{sarkar2020self}.
(3) Anomaly detection methods: SR \cite{ren2019time}, DONUT \cite{xu2018unsupervised}, SPOT and DSPOT \cite{siffer2017anomaly}.
(4) We also include GGS (Section \ref{sec:ggs}), which is a generally good strategy found by \autocss.

\textbf{Implementation.}
Most settings follow \cite{yue2022ts2vec}.
The embedding size of $f_C$ is 320, and MLPs are linear layers with softmax activations.




\subsection{Direct Application of \autocss}\label{sec:overall_performance}
In this subsection, we compare the proposed \autocss with existing CL methods to demonstrate the effectiveness of \autocss.
The overall performance of different methods on 7 public benchmark datasets and 3 downstream tasks are presented in Table \ref{tab:classification}-\ref{tab:forecast}.
Among all the baselines, InfoTS, which could automatically search suitable data augmentations, performs better than other baselines, which use handcrafted CLS.
This observation shows the effectiveness of automatically searching the data augmentations.
\autocss further outperforms InfoTS, indicating the overall effectiveness of the proposed search space and search algorithm.

\begin{table}[t]
    \centering
    \scriptsize
    \setlength\tabcolsep{3pt}
    \caption{Classification results on HAR and Epilepsy.}
    \begin{tabular}{c|c|cc|cccccccc}
    \toprule
    Datasets & Metrics & \autocss & GGS & InfoTS & TS2Vec & TS-TCC & CPC & Self-EEG\\
    \midrule
    \multirow{2}{*}{HAR} & ACC & \textbf{0.963} & \underline{0.937} & 0.930 & 0.930 & 0.904 & 0.839 & 0.653\\
    & F1 & \textbf{0.963} & \underline{0.937} & 0.929 & 0.930 & 0.904 & 0.833 & 0.638\\
    \midrule
    \multirow{2}{*}{Epilepsy} & ACC & \textbf{0.982} & \underline{0.977} & 0.976 & 0.975 & 0.972 & 0.966 & 0.937\\
    & F1 & \textbf{0.972} & \underline{0.963} & 0.962 & 0.962 & 0.955 & 0.944 & 0.892\\
    \bottomrule
    \end{tabular}
    \label{tab:classification}
    \vspace{-5pt}
\end{table}

\begin{table}[t]
    \centering
    \scriptsize
    \caption{Anomaly detection results on Yahoo and KPI.}
    \begin{tabular}{c|ccc|ccc}
    \toprule
    & \multicolumn{3}{c|}{Yahoo} & \multicolumn{3}{c}{KPI}\\
    \midrule
    Methods & F1 & Precision & Recall & F1 & Precision & Recall\\
    \midrule
    SPOT   & 0.338 & 0.269 & 0.454 & 0.217 & 0.786 & 0.126\\
    DSPOT  & 0.316 & 0.241 & 0.458 & 0.521 & 0.623 & 0.447\\
    DONUT  & 0.026 & 0.013 & 0.825 & 0.347 & 0.371 & 0.326\\
    SR     & 0.563 & 0.451 & 0.747 & 0.622 & 0.647 & 0.598\\
    TS2Vec & 0.745 & 0.729 & 0.762 & 0.677 & 0.929 & 0.533\\
    InfoTS & 0.746 & 0.744 & 0.747 & 0.672 & 0.927 & 0.526\\
    \midrule
    GGS & \underline{0.755} & 0.766 & 0.745 & \underline{0.689} & 0.807 & 0.601\\
    \autocss\ & \textbf{0.758} & 0.808 & 0.715 & \textbf{0.694} & 0.901 & 0.565\\
    \bottomrule
    \end{tabular}
    \label{tab:anomaly}
    \vspace{-10pt}
\end{table}

\begin{table*}[t]
    \centering
    \scriptsize
    \caption{Univariate time series forecasting results on ETTh1/ETTh2/ETTm1. The lower, the better.}
    \vspace{-10pt}
    \begin{tabular}{c|c|cccc|cccccccccc}
    \toprule
     & & \multicolumn{2}{c}{\autocss} & \multicolumn{2}{c|}{GGS} & \multicolumn{2}{c}{InfoTS} & \multicolumn{2}{c}{TS2Vec} & \multicolumn{2}{c}{CoST} & \multicolumn{2}{c}{TNC} & \multicolumn{2}{c}{TS-TCC} \\
    \cmidrule{3-16}
     Dataset & Horizon & MSE & MAE & MSE & MAE & MSE & MAE & MSE & MAE & MSE & MAE & MSE & MAE & MSE & MAE \\
    \midrule
    \multirow{5}{*}{ETTh1} & 24 & \textbf{0.036} & \textbf{0.142} & \underline{0.038} & \underline{0.146} & 0.039 & 0.149 & 0.039 & 0.151 & 0.040 & 0.152 & 0.057 & 0.184 & 0.103 & 0.237\\
    & 48  & \textbf{0.052} & \textbf{0.173} & \underline{0.054} & \underline{0.177} & 0.056 & 0.179 & 0.062 & 0.189 & 0.060 & 0.186 & 0.094 & 0.239 & 0.139 & 0.279\\
    & 168 & \textbf{0.087} & \textbf{0.223} & \underline{0.088} & \underline{0.227} & 0.100 & 0.239 & 0.142 & 0.291 & 0.097 & 0.236 & 0.171 & 0.329 & 0.253 & 0.408\\
    & 336 & \textbf{0.103} & \textbf{0.247} & \textbf{0.103} & \underline{0.248} & 0.117 & 0.264 & 0.160 & 0.316 & 0.112 & 0.258 & 0.192 & 0.357 & 0.155 & 0.318\\
    & 720 & \textbf{0.113} & \textbf{0.266} & \underline{0.123} & \underline{0.274} & 0.141 & 0.302 & 0.179 & 0.345 & 0.148 & 0.306 & 0.235 & 0.408 & 0.190 & 0.337\\
    \midrule
    \multirow{5}{*}{ETTh2} & 24 & \textbf{0.076} & \textbf{0.204} & \textbf{0.076} & \underline{0.205} & 0.081 & 0.215 & 0.091 & 0.230 & 0.079 & 0.207 & 0.097 & 0.238 & 0.239 & 0.391\\
    & 48 & \textbf{0.106} & \textbf{0.246} & \underline{0.107} & \underline{0.248} & 0.115 & 0.261 & 0.124 & 0.274 & 0.118 & 0.259 & 0.131 & 0.281 & 0.260 & 0.405\\
    & 168 & \textbf{0.169} & \textbf{0.314} & 0.172 & \underline{0.318} & \underline{0.171} & 0.327 & 0.198 & 0.355 & 0.189 & 0.339 & 0.197 & 0.354 & 0.291 & 0.420\\
    & 336 & \underline{0.184} & \underline{0.344} & 0.188 & 0.346 & \textbf{0.183} & \textbf{0.341} & 0.205 & 0.364 & 0.206 & 0.360 & 0.207 & 0.366 & 0.336 & 0.453\\
    & 720 & \textbf{0.189} & \textbf{0.354} & 0.196 & 0.359 & \underline{0.194} & \underline{0.357} & 0.208 & 0.371 & 0.214 & 0.371 & 0.207 & 0.370 & 0.362 & 0.472\\
    \midrule
    \multirow{5}{*}{ETTm1} & 24 & \textbf{0.013} & \textbf{0.084} & \underline{0.014} & \underline{0.085} & \underline{0.014} & 0.087 & 0.016 & 0.093 & 0.015 & 0.088 & 0.019 & 0.103 & 0.089 & 0.228\\
    & 48 & \textbf{0.024} & \textbf{0.114} & \underline{0.025} & \underline{0.117} & \underline{0.025} & \underline{0.117} & 0.028 & 0.126 & 0.025 & 0.117 & 0.036 & 0.142 & 0.134 & 0.280\\
    & 96 & \textbf{0.036} & \textbf{0.142} & \underline{0.038} & \underline{0.147} & \textbf{0.036} & \textbf{0.142} & 0.045 & 0.162 & 0.038 & 0.147 & 0.054 & 0.178 & 0.159 & 0.305\\
    & 288 & \textbf{0.071} & \underline{0.201} & \underline{0.078} & 0.211 & \textbf{0.071} & \textbf{0.200} & 0.095 & 0.235 & 0.077 & 0.209 & 0.098 & 0.244 & 0.204 & 0.327\\
    & 672 & \textbf{0.100} & \textbf{0.240} & 0.113 & \underline{0.255} & \underline{0.102} & \textbf{0.240} & 0.142 & 0.290 & 0.113 & 0.257 & 0.136 & 0.290 & 0.206 & 0.354\\
    \bottomrule
    \end{tabular}
    \label{tab:forecast}
\end{table*}

\begin{table*}[th!]
    \centering
    \setlength\tabcolsep{2.5pt}
    \scriptsize
    \caption{The Generally Good Strategy (GGS) derived by our \autocss method.}
    \vspace{-10pt}
    \begin{tabular}{c|c|c|c|c|c|c|c|c|c|c|c|c|c|c|c|c|c|c}
    \toprule
    \multicolumn{7}{c|}{Data Augmentations} & \multicolumn{3}{c|}{Embedding Transformations} & \multicolumn{6}{c|}{Contrastive Pair Construction} & \multicolumn{3}{c}{Loss Functions}\\
    \midrule
    Resize & Rescale & Jitter & Point Mask & Freq. Mask & Crop & Order & Emb. Jitter & Emb. Mask & Norm & Instance & Temporal & Cross-Scale & Kernel & Pool & Adj. Neighbor & Loss Type & Sim. Func. & Scaling \\
    \midrule
    0.2 & 0.3 & 0.0 & 0.2 & 0.0 & 0.2 & 3 & 0.7 & 0.1 & None & True & False & False & 5 & Avg & False & InfoNCE & Distance & 1.0\\
    \bottomrule
    \end{tabular}
    \label{tab:ggs}
    \vspace{-5pt}
\end{table*}

\subsection{Transferability Study}\label{sec:ggs}
Although \autocss could efficaciously discover the suitable CLS for given datasets/tasks, in many cases, searching could still be resource-consuming. 
Therefore, we seek to find a Generally Good Strategy (GGS) which can be used as a strong baseline and a good starting point for new datasets and tasks.

We search GGS on 3 datasets HAR/Yahoo/ETTh1, and then test it on all the datasets.
The procedure of finding GGS is described as follows.
{(1)} Collect sets of top $K$ candidate CLS from each of HAR/Yahoo/ETTh1, and use Cartesian product to obtain the $K^{3}$ pairs $(A_1, A_2, A_3)$.
{(2)} Obtain the top candidate pair by the number of the shared sub-dimensions across $A_1, A_2, A_3$, and add the shared options in GGS.
{(3)} For the un-shared sub-dimensions, discard the options that lead to a significant validation performance drop.

\begin{table}[t]
    \centering
    \scriptsize
    \setlength\tabcolsep{3pt}
    \caption{Ablation Study on HAR}
    \vspace{-10pt}
    \begin{tabular}{c|c|ccccc}
    \toprule
    Metrics & \autocss & Data Aug. Only & Full Pre-train $f_E$ & w/o Reward Filtering\\
    \midrule
    ACC & 0.963 & 0.942 & \textbf{0.965} & 0.930\\
    F1 & 0.963 & 0.941 & \textbf{0.964} &  0.931\\
    Time (hours) & 10.516 & 10.218 & 85.193 & 10.724\\
    \bottomrule
    \end{tabular}
    \label{tab:ablation}
\end{table}

A GGS we found is shown in Table \ref{tab:ggs}.
There are some interesting findings for the 4 dimensions.
For the \textbf{data augmentations}, small perturbations on the input time series are preferred. 
For the \textbf{embedding transformations}, large jitterring is preferred. 
Besides, as a compromise for the general good across different datasets and tasks, normalization is not adopted by GGS.
In fact, we have observed in Section \ref{sec:empirical_findings} that classification tasks could benefit from LayerNorm yet forecasting disfavors normalization.
For the \textbf{contrastive pair construction}, the instance contrast is the most important while the other aspects, i.e., the temporal and cross-scale contrasts, are less important.
For the \textbf{contrastive losses}, the popular InfoNCE loss is generally good for all datasets and tasks. 
For the similarity function, negative Euclidean distance is preferred over the popular dot product and cosine similarity.
Euclidean distance measures the distance between two points, yet cosine similarity and dot product also consider the angle between the two points.
This observation indicates that distance is more effective than angles for embeddings of time series data.
The performance of GGS on all datasets/tasks is presented in Table \ref{tab:classification}-\ref{tab:forecast}.
It can be observed that GGS outperforms all the baselines in general, but underperforms \autocss.

\subsection{Ablation Study}\label{sec:ablation_study}
In this subsection, we investigate the impact of different components of \autocss\ on HAR, and the results are shown in Table \ref{tab:ablation}.

\textbf{Effectiveness of the solution space.}
Compared with existing studies, which only consider the data augmentation dimension of CLS, our proposed solution space is more comprehensive, which is comprised of 4 most important dimensions: data augmentations, embedding transformations, contrastive pair construction and contrastive losses (Table \ref{tab:space}).
We can observe that the CLS obtained from the full space (\autocss) significantly outperforms the CLS obtained from the data augmentation dimension alone (Data Aug. Only).

\textbf{Effectiveness of the first-order approximation and the separate search/evaluation phases.}
We compare \autocss with the full pre-training of $f_E$ during phase 1, i.e., given the CLS $A$ sampled by $f_C$, we pre-train the encoder $f_E$ from scratch till the max number of iterations is reached.
The results in Table \ref{tab:ablation} show that the CLS found by the first-order approximation (\autocss) is competitive to the full pre-training of $f_E$ in terms of ACC and F1. 
However, \autocss is more than 8$\times$ faster than the full pre-training.
This is because \autocss:
(1) only pre-trains $f_E$ for one epoch for each iteration in phase 1;
(2) evaluates candidates in parallel in phase 2.

\begin{figure}[t]
\centering
\begin{subfigure}[b]{.22\textwidth}
\includegraphics[width=\linewidth]{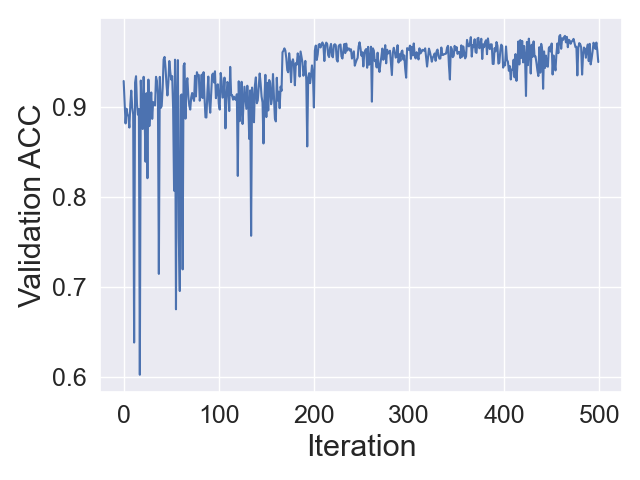}
 \caption{With reward filtering.}\label{fig:filter}
\end{subfigure}
\begin{subfigure}[b]{.22\textwidth}
  \includegraphics[width=\linewidth]{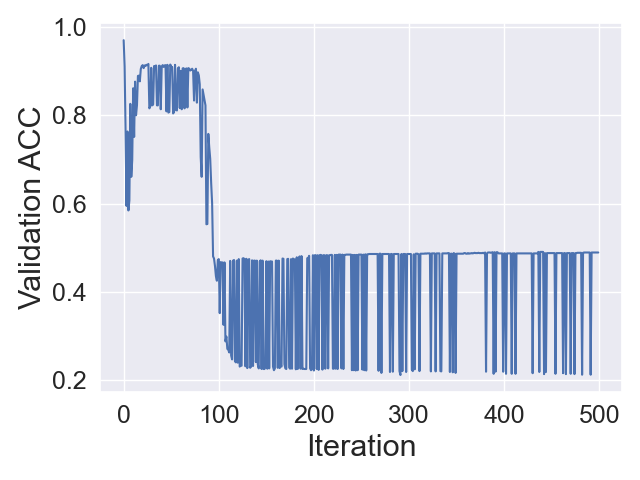}
  \caption{Without reward filtering.}\label{fig:no_filter}
\end{subfigure}
\vspace{-10pt}
\caption{Training curves of the models.} 
\vspace{-10pt}
\label{fig:training_curve}
\end{figure}

\textbf{Effectiveness of reward filtering.}
In Equation \eqref{eq:final_reward}, we use reward filtering to improve the process of RL.
In Table \ref{tab:ablation}, significant performance drop is observed if we remove the reward filtering.
In Figure \ref{fig:filter}, we can see that \autocss can progressively obtain higher and higher validation ACC.
However, as shown in Figure \ref{fig:no_filter}, if we remove reward filtering, the model fails quickly after the initial exploration in the first 100 steps.

\begin{figure*}[t]
\centering
\begin{subfigure}[b]{.16\textwidth}
\includegraphics[width=\linewidth]{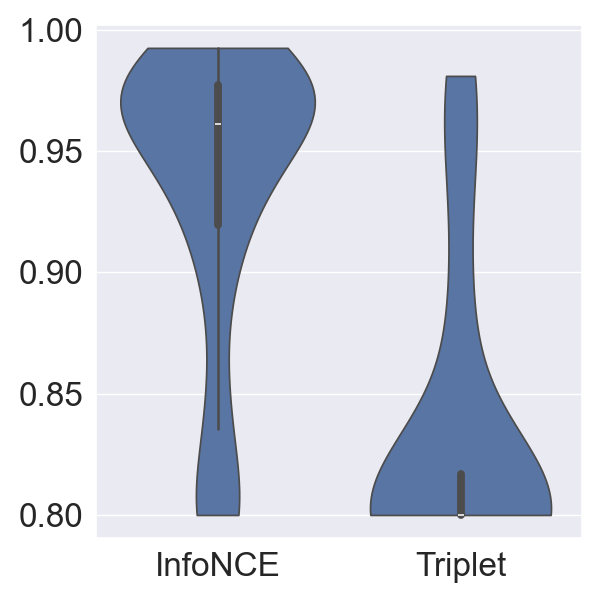}
 \caption{HAR: Loss Types}\label{fig:har_op}
\end{subfigure}
\begin{subfigure}[b]{.16\textwidth}
  \includegraphics[width=\linewidth]{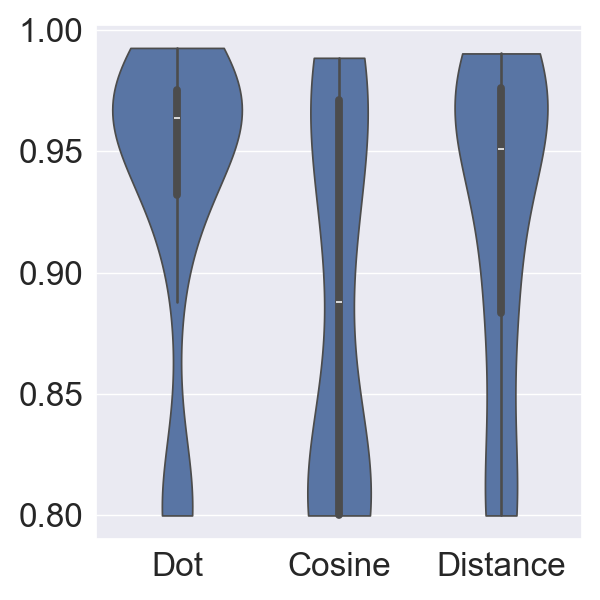}
  \caption{HAR: Sim. Functs.}\label{fig:har_sim}
\end{subfigure}
\begin{subfigure}[b]{.16\textwidth}
  \includegraphics[width=\linewidth]{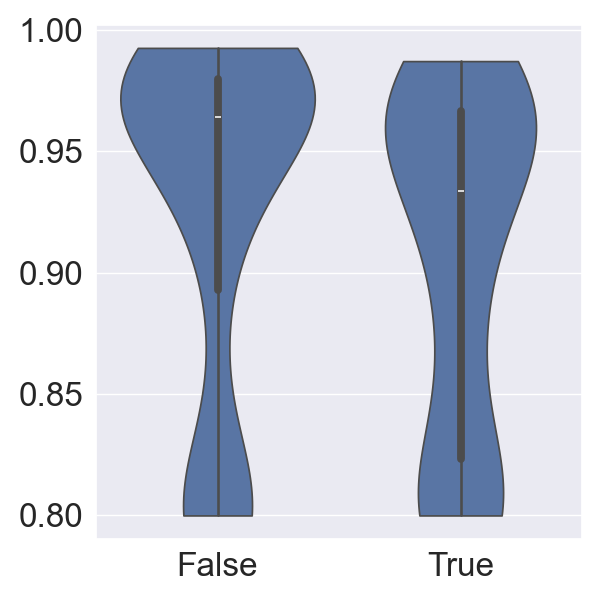}
  \caption{HAR: Temporal}\label{fig:har_w_t}
\end{subfigure}
\begin{subfigure}[b]{.16\textwidth}
\includegraphics[width=\linewidth]{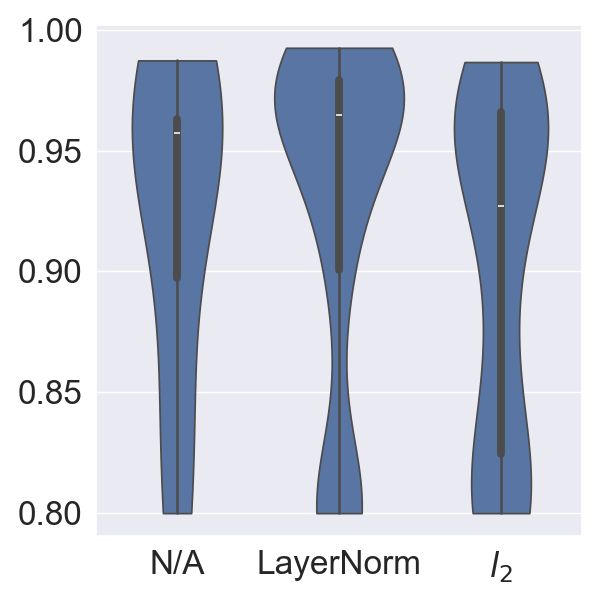}
 \caption{HAR: Norm}\label{fig:har_norm}
\end{subfigure}
\begin{subfigure}[b]{.16\textwidth}
  \includegraphics[width=\linewidth]{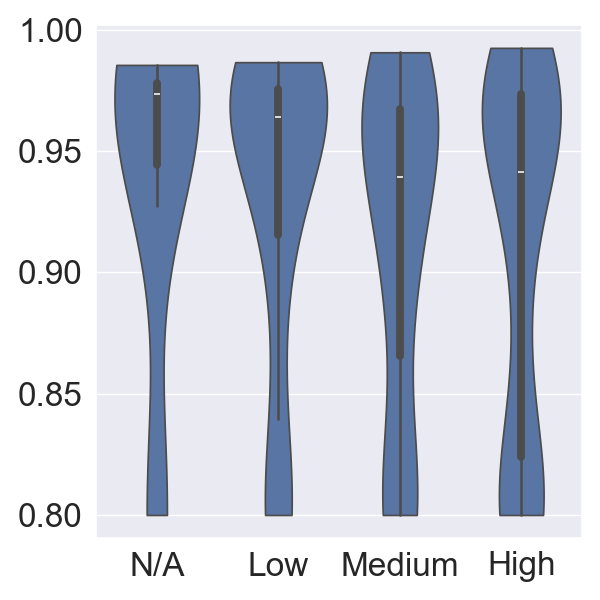}
  \caption{HAR: Emb. Jitter}\label{fig:har_jitter_emb}
\end{subfigure}
\begin{subfigure}[b]{.16\textwidth}
  \includegraphics[width=\linewidth]{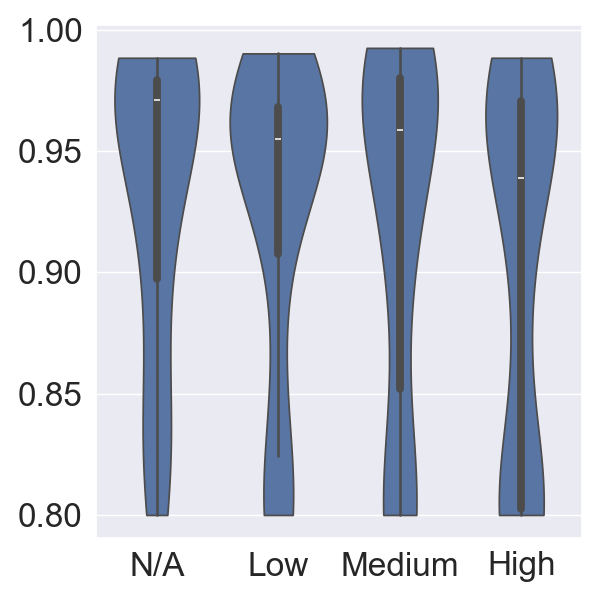}
  \caption{HAR: Freq. Mask}\label{fig:har_freq_mask}
\end{subfigure}

\begin{subfigure}[b]{.16\textwidth}
\includegraphics[width=\linewidth]{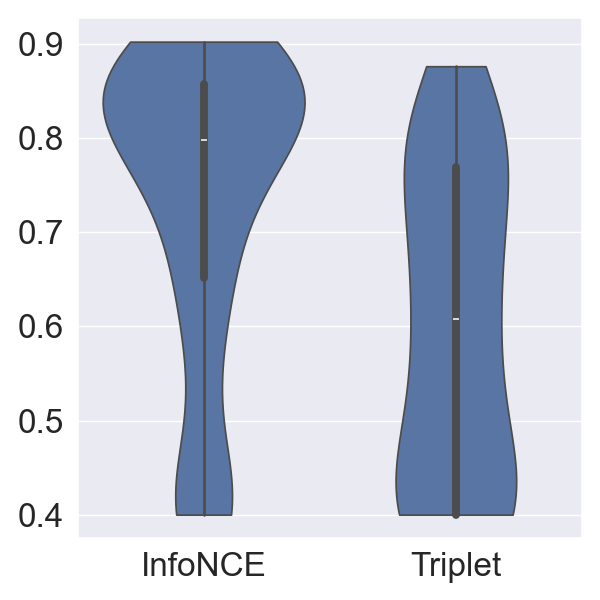}
 \caption{Yahoo: Loss Types}\label{fig:yahoo_op}
\end{subfigure}
\begin{subfigure}[b]{.16\textwidth}
  \includegraphics[width=\linewidth]{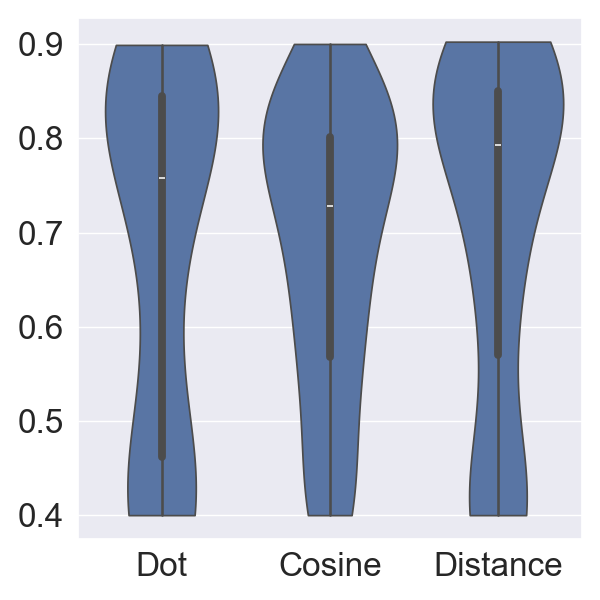}
  \caption{Yahoo: Sim. Functs.}\label{fig:yahoo_sim}
\end{subfigure}
\begin{subfigure}[b]{.16\textwidth}
  \includegraphics[width=\linewidth]{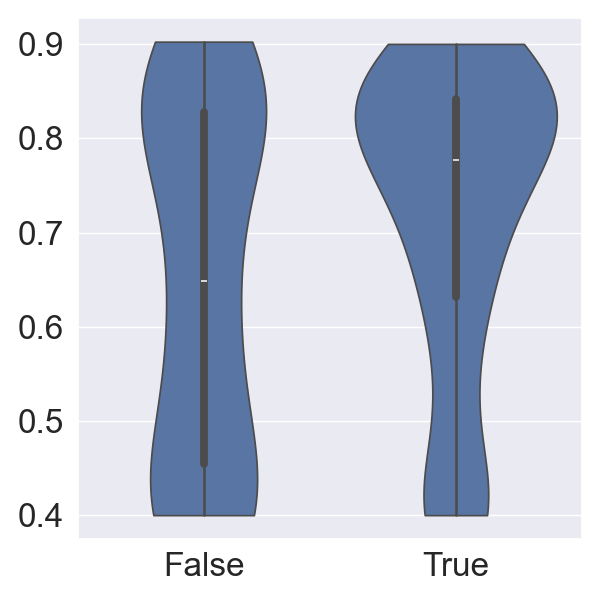}
  \caption{Yahoo: Temporal}\label{fig:yahoo_w_t}
\end{subfigure}
\begin{subfigure}[b]{.16\textwidth}
\includegraphics[width=\linewidth]{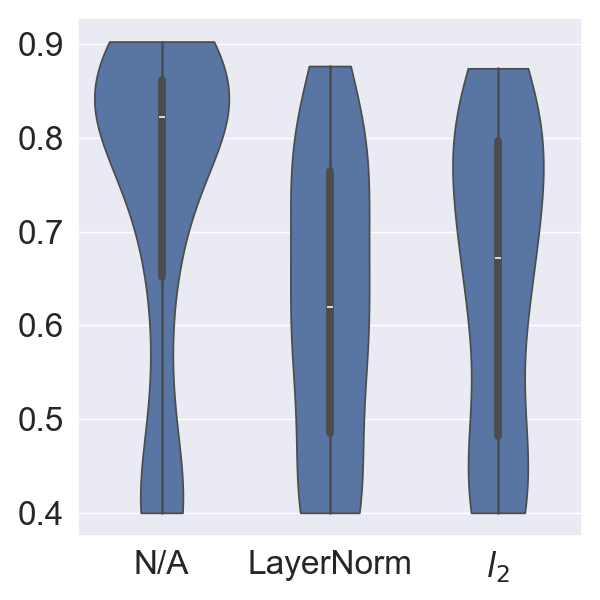}
 \caption{Yahoo: Norm}\label{fig:yahoo_norm}
\end{subfigure}
\begin{subfigure}[b]{.16\textwidth}
  \includegraphics[width=\linewidth]{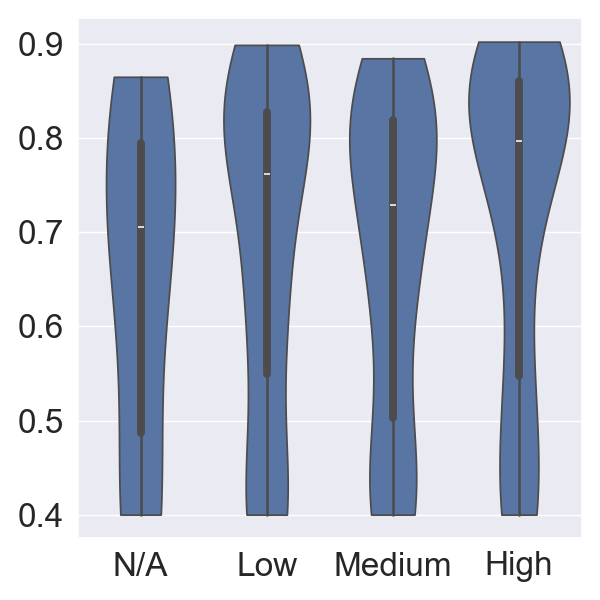}
  \caption{Yahoo: Emb. Jitter}\label{fig:yahoo_jitter_emb}
\end{subfigure}
\begin{subfigure}[b]{.16\textwidth}
  \includegraphics[width=\linewidth]{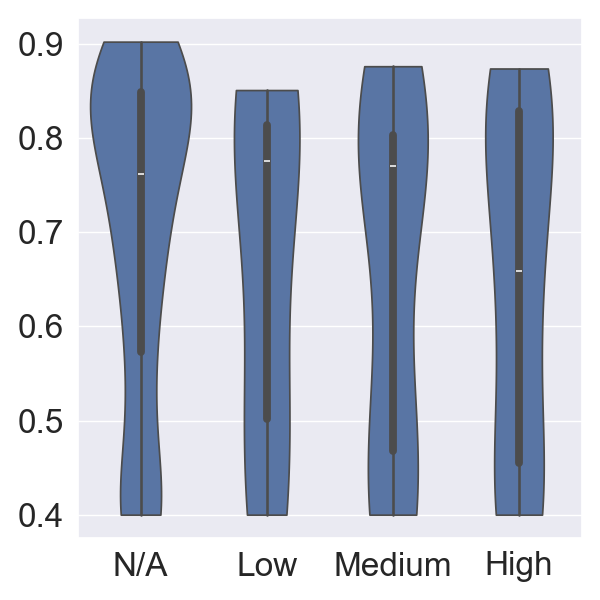}
  \caption{Yahoo: Freq. Mask}\label{fig:yahoo_freq}
\end{subfigure}

\begin{subfigure}[b]{.16\textwidth}
\includegraphics[width=\linewidth]{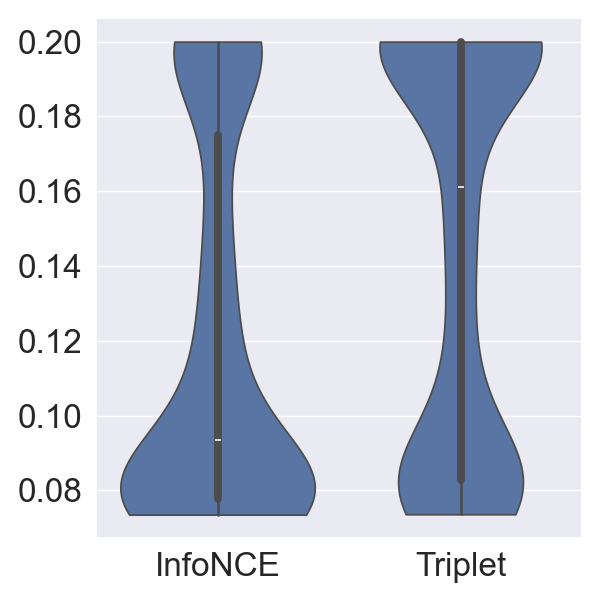}
 \caption{ETTh1: Loss Types}\label{fig:ett_op}
\end{subfigure}
\begin{subfigure}[b]{.16\textwidth}
  \includegraphics[width=\linewidth]{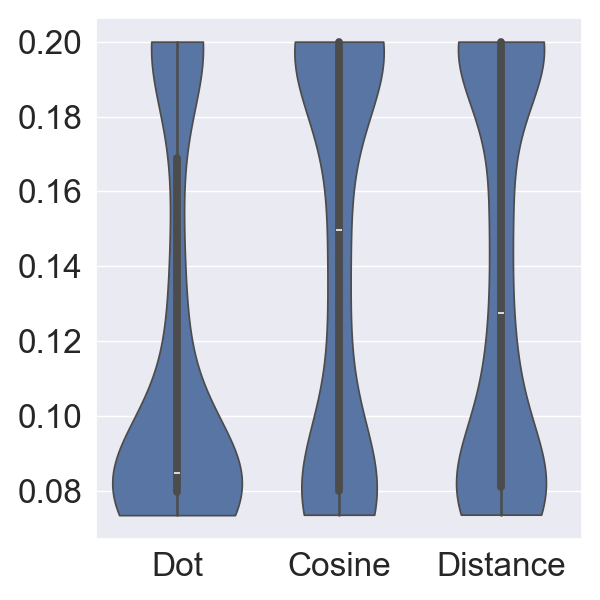}
  \caption{ETTh1: Sim. Functs.}\label{fig:ett_sim}
\end{subfigure}
\begin{subfigure}[b]{.16\textwidth}
  \includegraphics[width=\linewidth]{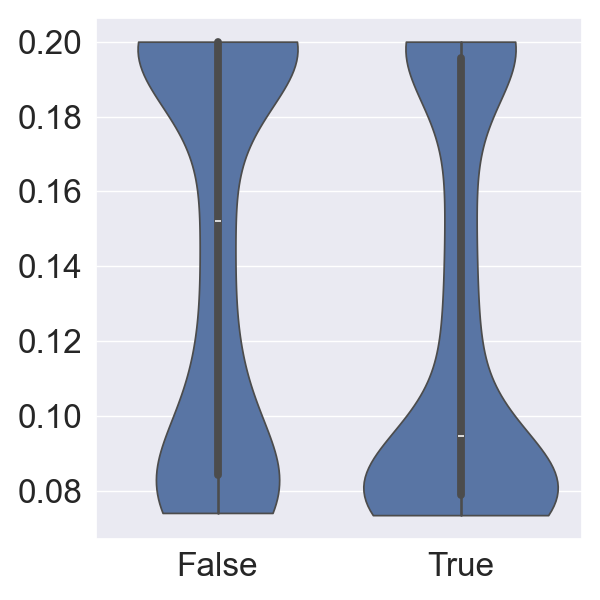}
  \caption{ETTh1: Temporal}\label{fig:ett_w_t}
\end{subfigure}
\begin{subfigure}[b]{.16\textwidth}
\includegraphics[width=\linewidth]{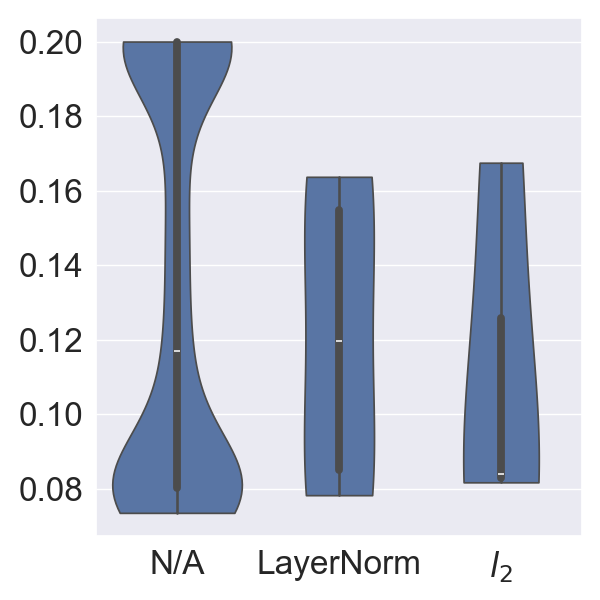}
 \caption{ETTh1: Norm}\label{fig:ett_norm}
\end{subfigure}
\begin{subfigure}[b]{.16\textwidth}
  \includegraphics[width=\linewidth]{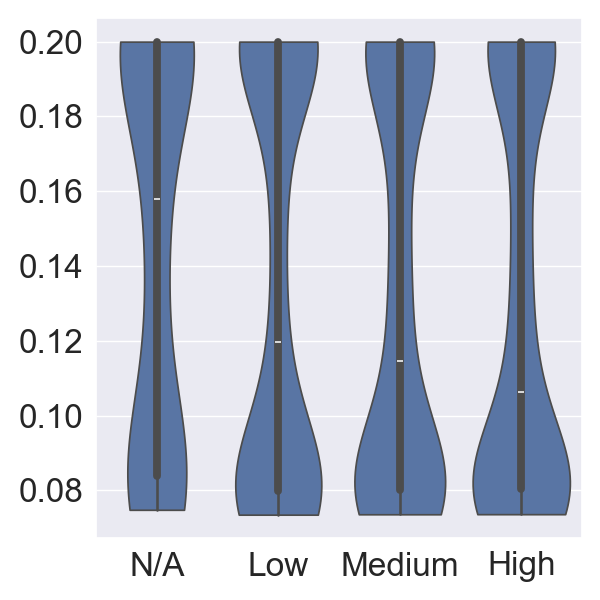}
  \caption{ETTh1: Emb. Jitter}\label{fig:ett_jitter_emb}
\end{subfigure}
\begin{subfigure}[b]{.16\textwidth}
  \includegraphics[width=\linewidth]{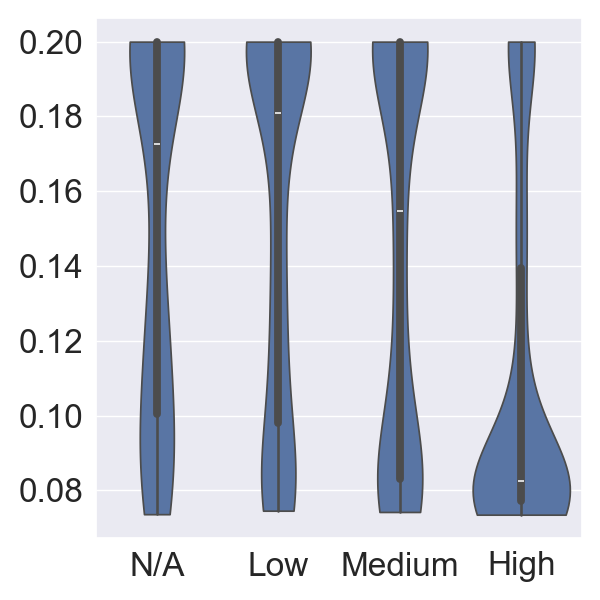}
  \caption{ETTh1: Freq. Mask}\label{fig:ett_freq}
\end{subfigure}

\caption{Violin plots of 6 sub-dimensions for HAR/Yahoo/ETTh1. X-axis: options. Y-axis: ACC/F1/MSE(48 horizon). ACC/F1: the higher the better. MSE: the lower the better. Low, Medium and High correspond to $0.1\sim0.3$, $0.4\sim0.6$ and $0.7\sim0.95$.} 
\label{fig:violin}
\end{figure*}

\subsection{Empirical Analysis of the Candidates}\label{sec:empirical_findings}
In this subsection, we provide in-depth empirical analysis based on the candidate CLS, and hope to provide potential insights and guidelines for future CLS designs.
Violin plots of 6 sub-dimensions for HAR/Yahoo/ETTh1 are shown in Figure \ref{fig:violin}, where rows correspond to datasets and columns correspond to sub-dimensions.

\textbf{Loss Types.}
The 1st column shows that InfoNCE is generally better than Triplet, indicating that comparing the similarities scores of the positive/negative pairs in the probability space usually yields better performance than comparing them in their original space.

\textbf{Similarity Functions.}
The 2nd column shows that dot product and negative Euclidean distance are generally better than cosine similarity. We believe this is because cosine similarity only captures the angle between two data points, yet ignores the magnitudes.

\textbf{Temporal Contrast.}
The 3rd column indicates that the temporal contrast is generally good for forecasting and anomaly detection, but it is less useful for classification. During pre-training, the input time series of ETTh1 and Yahoo are very long (2,000 steps) but the inputs of HAR are very short (128 steps).
This observation shows that the temporal contrast is more valuable for long time series.

\textbf{Normalization.}
The 4th column shows that LayerNorm is important for classification, however, the embedding normalization has a negative impact on forecasting and anomaly detection. 
We suspect this is because the normalization might destroy the fine-grained information, e.g., small anomaly points, which might be critical for forecasting and anomaly detection.

\textbf{Embedding Jittering.}
The 5th column indicates that applying jittering on embeddings is generally useful.
This finding is consistent with \cite{wang2019implicit} that jittering injects various small semantic perturbations into the embeddings, which helps the model capture the core semantics during contrastive pre-training.

\textbf{Frequency Masking.}
In the last column, Figure \ref{fig:ett_freq} shows that frequency masking with a high masking ratio could easily lead to good performance (low MSE).
We believe this is because frequency masking could remove the short-range noisy patterns, and make the model focus on the long-range patterns.
However, Figure \ref{fig:yahoo_freq} shows that frequency masking has a negative impact on anomaly detection.
Essentially, the anomaly points are sudden and short-range noise, which only accounts for a small ratio, e.g., 1\% of the entire time series.
Fourier Transform (FT) and Inverse Fourier Transform (IFT) involved in the frequency masking can easily remove these points from the input, and thus the model is unable to learn the information about anomaly points during pre-training.

\subsection{Experiments in Deployed Application}\label{sec:deployment}
We deploy \autocss to the machine learning platform of our partner, Shanghai Ruijin Hospital, a top-tier hospital in China.
The medical practitioners can easily utilize our \autocss to obtain effective time series models for analyzing various physiological time series data collected in medicine, e.g., EEG and ECG time series in healthcare.

We compare \autocss and GGS with strong baselines, i.e., InfoTS and TS2Vec, on detecting epileptic seizure from SEEG data.
These experiments, conducted with anonymized data, have received ethical approval from the hospital.
The results are shown in Table \ref{tab:classification_seeg}.
First, \autocss has the best overall performance, showing the superiority of \autocss in real-world applications.
Second, GGS is slightly better than the recent automatic data augmentation method InfoTS, indicating that GGS has a strong transferability and is a strong baseline for new datasets and tasks.
Consequently, in scenarios where a comprehensive search is prohibitively extensive, GGS presents a viable alternative. Moreover, empirical evidence from Section \ref{sec:empirical_findings} suggests that the incorporation of Layer Normalization typically enhances performance. 
Integrating Layer Normalization within the GGS, as evidenced by that in Table \ref{tab:classification_seeg}, yields performance improvements, thereby affirming the efficacy of this insight.

\begin{table}[t]
    \centering
    \scriptsize
    \caption{Classification results on the real-world application.}
    \begin{tabular}{c|ccc|cc}
    \toprule
    Metrics & \autocss & GGS+LayerNorm & GGS & InfoTS & TS2Vec\\
    \midrule
    ACC & \textbf{0.760} & 0.739 & 0.737 & 0.735 & 0.728 \\
    F1 & \textbf{0.756} & 0.734 & 0.730 & 0.730 & 0.719 \\
    \bottomrule
    \end{tabular}
    \vspace{-10pt}
    \label{tab:classification_seeg}
\end{table}

\section{Conclusion and Future Work}

In this study, we introduced an innovative framework, \autocss, for the automated discovery of Contrastive Learning Strategies (CLS) for time series, aimed at diminishing the reliance on extensive domain expertise and iterative experimentation in the development of CLS tailored to specific datasets and tasks. 
\autocss is structured around two principal components: a comprehensive solution space, meticulously designed to encapsulate four critical dimensions of CLS, and an effective optimization algorithm to identify suitable and transferable CLS configurations. 
Empirical evaluation on eight datasets demonstrates \autocss's effectiveness in finding suitable CLS, with the derived Generally Good Strategy (GGS) showing robust performance across different tasks. This study not only showcases the potential of \autocss but also provides valuable insights for future contrastive learning research.

\balance{
\bibliographystyle{ACM-Reference-Format}
\bibliography{sample-base}


\begin{thebibliography}{63}


\ifx \showCODEN    \undefined \def \showCODEN     #1{\unskip}     \fi
\ifx \showDOI      \undefined \def \showDOI       #1{#1}\fi
\ifx \showISBNx    \undefined \def \showISBNx     #1{\unskip}     \fi
\ifx \showISBNxiii \undefined \def \showISBNxiii  #1{\unskip}     \fi
\ifx \showISSN     \undefined \def \showISSN      #1{\unskip}     \fi
\ifx \showLCCN     \undefined \def \showLCCN      #1{\unskip}     \fi
\ifx \shownote     \undefined \def \shownote      #1{#1}          \fi
\ifx \showarticletitle \undefined \def \showarticletitle #1{#1}   \fi
\ifx \showURL      \undefined \def \showURL       {\relax}        \fi
\providecommand\bibfield[2]{#2}
\providecommand\bibinfo[2]{#2}
\providecommand\natexlab[1]{#1}
\providecommand\showeprint[2][]{arXiv:#2}

\bibitem[Andrzejak et~al\mbox{.}(2001)]%
        {andrzejak2001indications}
\bibfield{author}{\bibinfo{person}{Ralph~G Andrzejak}, \bibinfo{person}{Klaus Lehnertz}, \bibinfo{person}{Florian Mormann}, \bibinfo{person}{Christoph Rieke}, \bibinfo{person}{Peter David}, {and} \bibinfo{person}{Christian~E Elger}.} \bibinfo{year}{2001}\natexlab{}.
\newblock \showarticletitle{Indications of nonlinear deterministic and finite-dimensional structures in time series of brain electrical activity: Dependence on recording region and brain state}.
\newblock \bibinfo{journal}{\emph{Physical Review E}} \bibinfo{volume}{64}, \bibinfo{number}{6} (\bibinfo{year}{2001}), \bibinfo{pages}{061907}.
\newblock


\bibitem[Anguita et~al\mbox{.}(2013)]%
        {anguita2013public}
\bibfield{author}{\bibinfo{person}{Davide Anguita}, \bibinfo{person}{Alessandro Ghio}, \bibinfo{person}{Luca Oneto}, \bibinfo{person}{Xavier Parra}, \bibinfo{person}{Jorge~Luis Reyes-Ortiz}, {et~al\mbox{.}}} \bibinfo{year}{2013}\natexlab{}.
\newblock \showarticletitle{A public domain dataset for human activity recognition using smartphones.}. In \bibinfo{booktitle}{\emph{Esann}}, Vol.~\bibinfo{volume}{3}. \bibinfo{pages}{3}.
\newblock


\bibitem[Ba et~al\mbox{.}(2016)]%
        {ba2016layer}
\bibfield{author}{\bibinfo{person}{Jimmy~Lei Ba}, \bibinfo{person}{Jamie~Ryan Kiros}, {and} \bibinfo{person}{Geoffrey~E Hinton}.} \bibinfo{year}{2016}\natexlab{}.
\newblock \showarticletitle{Layer normalization}.
\newblock \bibinfo{journal}{\emph{arXiv preprint arXiv:1607.06450}} (\bibinfo{year}{2016}).
\newblock


\bibitem[Bachman et~al\mbox{.}(2019)]%
        {bachman2019learning}
\bibfield{author}{\bibinfo{person}{Philip Bachman}, \bibinfo{person}{R~Devon Hjelm}, {and} \bibinfo{person}{William Buchwalter}.} \bibinfo{year}{2019}\natexlab{}.
\newblock \showarticletitle{Learning representations by maximizing mutual information across views}.
\newblock \bibinfo{journal}{\emph{Advances in neural information processing systems}}  \bibinfo{volume}{32} (\bibinfo{year}{2019}).
\newblock


\bibitem[Beirlant et~al\mbox{.}(2006)]%
        {beirlant2006statistics}
\bibfield{author}{\bibinfo{person}{Jan Beirlant}, \bibinfo{person}{Yuri Goegebeur}, \bibinfo{person}{Johan Segers}, {and} \bibinfo{person}{Jozef~L Teugels}.} \bibinfo{year}{2006}\natexlab{}.
\newblock \bibinfo{booktitle}{\emph{Statistics of extremes: theory and applications}}.
\newblock \bibinfo{publisher}{John Wiley \& Sons}.
\newblock


\bibitem[Chen et~al\mbox{.}(2020)]%
        {chen2020simple}
\bibfield{author}{\bibinfo{person}{Ting Chen}, \bibinfo{person}{Simon Kornblith}, \bibinfo{person}{Mohammad Norouzi}, {and} \bibinfo{person}{Geoffrey Hinton}.} \bibinfo{year}{2020}\natexlab{}.
\newblock \showarticletitle{A simple framework for contrastive learning of visual representations}. In \bibinfo{booktitle}{\emph{International conference on machine learning}}. PMLR, \bibinfo{pages}{1597--1607}.
\newblock


\bibitem[Chen et~al\mbox{.}(2024a)]%
        {chen2024eegformer}
\bibfield{author}{\bibinfo{person}{Yuqi Chen}, \bibinfo{person}{Kan Ren}, \bibinfo{person}{Kaitao Song}, \bibinfo{person}{Yansen Wang}, \bibinfo{person}{Yifan Wang}, \bibinfo{person}{Dongsheng Li}, {and} \bibinfo{person}{Lili Qiu}.} \bibinfo{year}{2024}\natexlab{a}.
\newblock \showarticletitle{EEGFormer: Towards Transferable and Interpretable Large-Scale EEG Foundation Model}. In \bibinfo{booktitle}{\emph{AAAI 2024 Spring Symposium on Clinical Foundation Models}}.
\newblock


\bibitem[Chen et~al\mbox{.}(2024b)]%
        {chen2024contiformer}
\bibfield{author}{\bibinfo{person}{Yuqi Chen}, \bibinfo{person}{Kan Ren}, \bibinfo{person}{Yansen Wang}, \bibinfo{person}{Yuchen Fang}, \bibinfo{person}{Weiwei Sun}, {and} \bibinfo{person}{Dongsheng Li}.} \bibinfo{year}{2024}\natexlab{b}.
\newblock \showarticletitle{Contiformer: Continuous-time transformer for irregular time series modeling}.
\newblock \bibinfo{journal}{\emph{Advances in Neural Information Processing Systems}}  \bibinfo{volume}{36} (\bibinfo{year}{2024}).
\newblock


\bibitem[Cubuk et~al\mbox{.}(2019)]%
        {cubuk2019autoaugment}
\bibfield{author}{\bibinfo{person}{Ekin~D Cubuk}, \bibinfo{person}{Barret Zoph}, \bibinfo{person}{Dandelion Mane}, \bibinfo{person}{Vijay Vasudevan}, {and} \bibinfo{person}{Quoc~V Le}.} \bibinfo{year}{2019}\natexlab{}.
\newblock \showarticletitle{Autoaugment: Learning augmentation strategies from data}. In \bibinfo{booktitle}{\emph{Proceedings of the IEEE/CVF conference on computer vision and pattern recognition}}. \bibinfo{pages}{113--123}.
\newblock


\bibitem[Doersch(2016)]%
        {doersch2016tutorial}
\bibfield{author}{\bibinfo{person}{Carl Doersch}.} \bibinfo{year}{2016}\natexlab{}.
\newblock \showarticletitle{Tutorial on variational autoencoders}.
\newblock \bibinfo{journal}{\emph{arXiv preprint arXiv:1606.05908}} (\bibinfo{year}{2016}).
\newblock


\bibitem[Eldele et~al\mbox{.}(2021)]%
        {eldeletime}
\bibfield{author}{\bibinfo{person}{Emadeldeen Eldele}, \bibinfo{person}{Mohamed Ragab}, \bibinfo{person}{Zhenghua Chen}, \bibinfo{person}{Min Wu}, \bibinfo{person}{Chee~Keong Kwoh}, \bibinfo{person}{Xiaoli Li}, {and} \bibinfo{person}{Cuntai Guan}.} \bibinfo{year}{2021}\natexlab{}.
\newblock \showarticletitle{Time-Series Representation Learning via Temporal and Contextual Contrasting}. In \bibinfo{booktitle}{\emph{Proceedings of the Thirtieth International Joint Conference on Artificial Intelligence, {IJCAI-21}}}. \bibinfo{pages}{2352--2359}.
\newblock


\bibitem[Feng et~al\mbox{.}(2022)]%
        {feng2022adversarial}
\bibfield{author}{\bibinfo{person}{Shengyu Feng}, \bibinfo{person}{Baoyu Jing}, \bibinfo{person}{Yada Zhu}, {and} \bibinfo{person}{Hanghang Tong}.} \bibinfo{year}{2022}\natexlab{}.
\newblock \showarticletitle{Adversarial graph contrastive learning with information regularization}. In \bibinfo{booktitle}{\emph{Proceedings of the ACM Web Conference 2022}}. \bibinfo{pages}{1362--1371}.
\newblock


\bibitem[Feng et~al\mbox{.}(2024)]%
        {feng2024ariel}
\bibfield{author}{\bibinfo{person}{Shengyu Feng}, \bibinfo{person}{Baoyu Jing}, \bibinfo{person}{Yada Zhu}, {and} \bibinfo{person}{Hanghang Tong}.} \bibinfo{year}{2024}\natexlab{}.
\newblock \showarticletitle{Ariel: Adversarial graph contrastive learning}.
\newblock \bibinfo{journal}{\emph{ACM Transactions on Knowledge Discovery from Data}} \bibinfo{volume}{18}, \bibinfo{number}{4} (\bibinfo{year}{2024}), \bibinfo{pages}{1--22}.
\newblock


\bibitem[Franceschi et~al\mbox{.}(2019)]%
        {franceschi2019unsupervised}
\bibfield{author}{\bibinfo{person}{Jean-Yves Franceschi}, \bibinfo{person}{Aymeric Dieuleveut}, {and} \bibinfo{person}{Martin Jaggi}.} \bibinfo{year}{2019}\natexlab{}.
\newblock \showarticletitle{Unsupervised scalable representation learning for multivariate time series}.
\newblock \bibinfo{journal}{\emph{Advances in neural information processing systems}}  \bibinfo{volume}{32} (\bibinfo{year}{2019}).
\newblock


\bibitem[Han et~al\mbox{.}(2021)]%
        {han2021semi}
\bibfield{author}{\bibinfo{person}{Jinpei Han}, \bibinfo{person}{Xiao Gu}, {and} \bibinfo{person}{Benny Lo}.} \bibinfo{year}{2021}\natexlab{}.
\newblock \showarticletitle{Semi-supervised contrastive learning for generalizable motor imagery eeg classification}. In \bibinfo{booktitle}{\emph{2021 IEEE 17th International Conference on Wearable and Implantable Body Sensor Networks (BSN)}}. IEEE, \bibinfo{pages}{1--4}.
\newblock


\bibitem[He et~al\mbox{.}(2021)]%
        {he2021automl}
\bibfield{author}{\bibinfo{person}{Xin He}, \bibinfo{person}{Kaiyong Zhao}, {and} \bibinfo{person}{Xiaowen Chu}.} \bibinfo{year}{2021}\natexlab{}.
\newblock \showarticletitle{AutoML: A survey of the state-of-the-art}.
\newblock \bibinfo{journal}{\emph{Knowledge-Based Systems}}  \bibinfo{volume}{212} (\bibinfo{year}{2021}), \bibinfo{pages}{106622}.
\newblock


\bibitem[Hoffer and Ailon(2015)]%
        {hoffer2015deep}
\bibfield{author}{\bibinfo{person}{Elad Hoffer} {and} \bibinfo{person}{Nir Ailon}.} \bibinfo{year}{2015}\natexlab{}.
\newblock \showarticletitle{Deep metric learning using triplet network}. In \bibinfo{booktitle}{\emph{Similarity-Based Pattern Recognition: Third International Workshop, SIMBAD 2015, Copenhagen, Denmark, October 12-14, 2015. Proceedings 3}}. Springer, \bibinfo{pages}{84--92}.
\newblock


\bibitem[Jaiswal et~al\mbox{.}(2020)]%
        {jaiswal2020survey}
\bibfield{author}{\bibinfo{person}{Ashish Jaiswal}, \bibinfo{person}{Ashwin~Ramesh Babu}, \bibinfo{person}{Mohammad~Zaki Zadeh}, \bibinfo{person}{Debapriya Banerjee}, {and} \bibinfo{person}{Fillia Makedon}.} \bibinfo{year}{2020}\natexlab{}.
\newblock \showarticletitle{A survey on contrastive self-supervised learning}.
\newblock \bibinfo{journal}{\emph{Technologies}} \bibinfo{volume}{9}, \bibinfo{number}{1} (\bibinfo{year}{2020}), \bibinfo{pages}{2}.
\newblock


\bibitem[Jing et~al\mbox{.}(2022a)]%
        {jing2022x}
\bibfield{author}{\bibinfo{person}{Baoyu Jing}, \bibinfo{person}{Shengyu Feng}, \bibinfo{person}{Yuejia Xiang}, \bibinfo{person}{Xi Chen}, \bibinfo{person}{Yu Chen}, {and} \bibinfo{person}{Hanghang Tong}.} \bibinfo{year}{2022}\natexlab{a}.
\newblock \showarticletitle{X-GOAL: Multiplex heterogeneous graph prototypical contrastive learning}. In \bibinfo{booktitle}{\emph{Proceedings of the 31st ACM International Conference on Information \& Knowledge Management}}. \bibinfo{pages}{894--904}.
\newblock


\bibitem[Jing et~al\mbox{.}(2021a)]%
        {jing2021hdmi}
\bibfield{author}{\bibinfo{person}{Baoyu Jing}, \bibinfo{person}{Chanyoung Park}, {and} \bibinfo{person}{Hanghang Tong}.} \bibinfo{year}{2021}\natexlab{a}.
\newblock \showarticletitle{Hdmi: High-order deep multiplex infomax}. In \bibinfo{booktitle}{\emph{Proceedings of the Web Conference 2021}}. \bibinfo{pages}{2414--2424}.
\newblock


\bibitem[Jing et~al\mbox{.}(2021b)]%
        {jing2021network}
\bibfield{author}{\bibinfo{person}{Baoyu Jing}, \bibinfo{person}{Hanghang Tong}, {and} \bibinfo{person}{Yada Zhu}.} \bibinfo{year}{2021}\natexlab{b}.
\newblock \showarticletitle{Network of tensor time series}. In \bibinfo{booktitle}{\emph{Proceedings of the Web Conference 2021}}. \bibinfo{pages}{2425--2437}.
\newblock


\bibitem[Jing et~al\mbox{.}(2024a)]%
        {jing2024sterling}
\bibfield{author}{\bibinfo{person}{Baoyu Jing}, \bibinfo{person}{Yuchen Yan}, \bibinfo{person}{Kaize Ding}, \bibinfo{person}{Chanyoung Park}, \bibinfo{person}{Yada Zhu}, \bibinfo{person}{Huan Liu}, {and} \bibinfo{person}{Hanghang Tong}.} \bibinfo{year}{2024}\natexlab{a}.
\newblock \showarticletitle{Sterling: Synergistic representation learning on bipartite graphs}. In \bibinfo{booktitle}{\emph{Proceedings of the AAAI Conference on Artificial Intelligence}}, Vol.~\bibinfo{volume}{38}. \bibinfo{pages}{12976--12984}.
\newblock


\bibitem[Jing et~al\mbox{.}(2022b)]%
        {jing2022coin}
\bibfield{author}{\bibinfo{person}{Baoyu Jing}, \bibinfo{person}{Yuchen Yan}, \bibinfo{person}{Yada Zhu}, {and} \bibinfo{person}{Hanghang Tong}.} \bibinfo{year}{2022}\natexlab{b}.
\newblock \showarticletitle{Coin: Co-cluster infomax for bipartite graphs}. In \bibinfo{booktitle}{\emph{NeurIPS 2022 Workshop: New Frontiers in Graph Learning}}.
\newblock


\bibitem[Jing et~al\mbox{.}(2022c)]%
        {jing2022retrieval}
\bibfield{author}{\bibinfo{person}{Baoyu Jing}, \bibinfo{person}{Si Zhang}, \bibinfo{person}{Yada Zhu}, \bibinfo{person}{Bin Peng}, \bibinfo{person}{Kaiyu Guan}, \bibinfo{person}{Andrew Margenot}, {and} \bibinfo{person}{Hanghang Tong}.} \bibinfo{year}{2022}\natexlab{c}.
\newblock \showarticletitle{Retrieval based time series forecasting}.
\newblock \bibinfo{journal}{\emph{arXiv preprint arXiv:2209.13525}} (\bibinfo{year}{2022}).
\newblock


\bibitem[Jing et~al\mbox{.}(2024b)]%
        {jing2024casper}
\bibfield{author}{\bibinfo{person}{Baoyu Jing}, \bibinfo{person}{Dawei Zhou}, \bibinfo{person}{Kan Ren}, {and} \bibinfo{person}{Carl Yang}.} \bibinfo{year}{2024}\natexlab{b}.
\newblock \showarticletitle{CASPER: Causality-Aware Spatiotemporal Graph Neural Networks for Spatiotemporal Time Series Imputation}.
\newblock \bibinfo{journal}{\emph{arXiv preprint arXiv:2403.11960}} (\bibinfo{year}{2024}).
\newblock


\bibitem[Laptev et~al\mbox{.}(2015)]%
        {laptev2015benchmark}
\bibfield{author}{\bibinfo{person}{Nikolay Laptev}, \bibinfo{person}{Saeed Amizadeh}, {and} \bibinfo{person}{Youssef Billawala}.} \bibinfo{year}{2015}\natexlab{}.
\newblock \bibinfo{title}{A Benchmark Dataset for Time Series Anomaly Detection}.
\newblock
\newblock


\bibitem[Li et~al\mbox{.}(2022b)]%
        {li2022graph}
\bibfield{author}{\bibinfo{person}{Bolian Li}, \bibinfo{person}{Baoyu Jing}, {and} \bibinfo{person}{Hanghang Tong}.} \bibinfo{year}{2022}\natexlab{b}.
\newblock \showarticletitle{Graph communal contrastive learning}. In \bibinfo{booktitle}{\emph{Proceedings of the ACM web conference 2022}}. \bibinfo{pages}{1203--1213}.
\newblock


\bibitem[Li et~al\mbox{.}(2021)]%
        {li2021outlier}
\bibfield{author}{\bibinfo{person}{Jianbo Li}, \bibinfo{person}{Lecheng Zheng}, \bibinfo{person}{Yada Zhu}, {and} \bibinfo{person}{Jingrui He}.} \bibinfo{year}{2021}\natexlab{}.
\newblock \showarticletitle{Outlier impact characterization for time series data}. In \bibinfo{booktitle}{\emph{Proceedings of the AAAI Conference on Artificial Intelligence}}, Vol.~\bibinfo{volume}{35}. \bibinfo{pages}{11595--11603}.
\newblock


\bibitem[Li et~al\mbox{.}(2022a)]%
        {li2022autolossgen}
\bibfield{author}{\bibinfo{person}{Zelong Li}, \bibinfo{person}{Jianchao Ji}, \bibinfo{person}{Yingqiang Ge}, {and} \bibinfo{person}{Yongfeng Zhang}.} \bibinfo{year}{2022}\natexlab{a}.
\newblock \showarticletitle{AutoLossGen: Automatic loss function generation for recommender systems}. In \bibinfo{booktitle}{\emph{Proceedings of the 45th International ACM SIGIR Conference on Research and Development in Information Retrieval}}. \bibinfo{pages}{1304--1315}.
\newblock


\bibitem[Liu et~al\mbox{.}(2018)]%
        {liu2018darts}
\bibfield{author}{\bibinfo{person}{Hanxiao Liu}, \bibinfo{person}{Karen Simonyan}, {and} \bibinfo{person}{Yiming Yang}.} \bibinfo{year}{2018}\natexlab{}.
\newblock \showarticletitle{DARTS: Differentiable Architecture Search}. In \bibinfo{booktitle}{\emph{International Conference on Learning Representations}}.
\newblock


\bibitem[Liu and Chen(2024)]%
        {liu2024timesurl}
\bibfield{author}{\bibinfo{person}{Jiexi Liu} {and} \bibinfo{person}{Songcan Chen}.} \bibinfo{year}{2024}\natexlab{}.
\newblock \showarticletitle{Timesurl: Self-supervised contrastive learning for universal time series representation learning}. In \bibinfo{booktitle}{\emph{Proceedings of the AAAI Conference on Artificial Intelligence}}, Vol.~\bibinfo{volume}{38}. \bibinfo{pages}{13918--13926}.
\newblock


\bibitem[Liu et~al\mbox{.}(2021)]%
        {liu2021self}
\bibfield{author}{\bibinfo{person}{Xiao Liu}, \bibinfo{person}{Fanjin Zhang}, \bibinfo{person}{Zhenyu Hou}, \bibinfo{person}{Li Mian}, \bibinfo{person}{Zhaoyu Wang}, \bibinfo{person}{Jing Zhang}, {and} \bibinfo{person}{Jie Tang}.} \bibinfo{year}{2021}\natexlab{}.
\newblock \showarticletitle{Self-supervised learning: Generative or contrastive}.
\newblock \bibinfo{journal}{\emph{IEEE transactions on knowledge and data engineering}} \bibinfo{volume}{35}, \bibinfo{number}{1} (\bibinfo{year}{2021}), \bibinfo{pages}{857--876}.
\newblock


\bibitem[Liu et~al\mbox{.}(2023)]%
        {liu2023self}
\bibfield{author}{\bibinfo{person}{Ziyu Liu}, \bibinfo{person}{Azadeh Alavi}, \bibinfo{person}{Minyi Li}, {and} \bibinfo{person}{Xiang Zhang}.} \bibinfo{year}{2023}\natexlab{}.
\newblock \showarticletitle{Self-Supervised Contrastive Learning for Medical Time Series: A Systematic Review}.
\newblock \bibinfo{journal}{\emph{Sensors}} \bibinfo{volume}{23}, \bibinfo{number}{9} (\bibinfo{year}{2023}), \bibinfo{pages}{4221}.
\newblock


\bibitem[Luo et~al\mbox{.}(2023)]%
        {luo2023time}
\bibfield{author}{\bibinfo{person}{Dongsheng Luo}, \bibinfo{person}{Wei Cheng}, \bibinfo{person}{Yingheng Wang}, \bibinfo{person}{Dongkuan Xu}, \bibinfo{person}{Jingchao Ni}, \bibinfo{person}{Wenchao Yu}, \bibinfo{person}{Xuchao Zhang}, \bibinfo{person}{Yanchi Liu}, \bibinfo{person}{Yuncong Chen}, \bibinfo{person}{Haifeng Chen}, {et~al\mbox{.}}} \bibinfo{year}{2023}\natexlab{}.
\newblock \showarticletitle{Time series contrastive learning with information-aware augmentations}. In \bibinfo{booktitle}{\emph{Proceedings of the AAAI Conference on Artificial Intelligence}}, Vol.~\bibinfo{volume}{37}. \bibinfo{pages}{4534--4542}.
\newblock


\bibitem[Mehari and Strodthoff(2022)]%
        {mehari2022self}
\bibfield{author}{\bibinfo{person}{Temesgen Mehari} {and} \bibinfo{person}{Nils Strodthoff}.} \bibinfo{year}{2022}\natexlab{}.
\newblock \showarticletitle{Self-supervised representation learning from 12-lead ECG data}.
\newblock \bibinfo{journal}{\emph{Computers in biology and medicine}}  \bibinfo{volume}{141} (\bibinfo{year}{2022}), \bibinfo{pages}{105114}.
\newblock


\bibitem[Nie et~al\mbox{.}(2022)]%
        {nie2022time}
\bibfield{author}{\bibinfo{person}{Yuqi Nie}, \bibinfo{person}{Nam~H Nguyen}, \bibinfo{person}{Phanwadee Sinthong}, {and} \bibinfo{person}{Jayant Kalagnanam}.} \bibinfo{year}{2022}\natexlab{}.
\newblock \showarticletitle{A time series is worth 64 words: Long-term forecasting with transformers}.
\newblock \bibinfo{journal}{\emph{arXiv preprint arXiv:2211.14730}} (\bibinfo{year}{2022}).
\newblock


\bibitem[Oord et~al\mbox{.}(2018)]%
        {oord2018representation}
\bibfield{author}{\bibinfo{person}{Aaron van~den Oord}, \bibinfo{person}{Yazhe Li}, {and} \bibinfo{person}{Oriol Vinyals}.} \bibinfo{year}{2018}\natexlab{}.
\newblock \showarticletitle{Representation learning with contrastive predictive coding}.
\newblock \bibinfo{journal}{\emph{arXiv preprint arXiv:1807.03748}} (\bibinfo{year}{2018}).
\newblock


\bibitem[Ren et~al\mbox{.}(2019)]%
        {ren2019time}
\bibfield{author}{\bibinfo{person}{Hansheng Ren}, \bibinfo{person}{Bixiong Xu}, \bibinfo{person}{Yujing Wang}, \bibinfo{person}{Chao Yi}, \bibinfo{person}{Congrui Huang}, \bibinfo{person}{Xiaoyu Kou}, \bibinfo{person}{Tony Xing}, \bibinfo{person}{Mao Yang}, \bibinfo{person}{Jie Tong}, {and} \bibinfo{person}{Qi Zhang}.} \bibinfo{year}{2019}\natexlab{}.
\newblock \showarticletitle{Time-series anomaly detection service at microsoft}. In \bibinfo{booktitle}{\emph{Proceedings of the 25th ACM SIGKDD international conference on knowledge discovery \& data mining}}. \bibinfo{pages}{3009--3017}.
\newblock


\bibitem[Sarkar and Etemad(2020)]%
        {sarkar2020self}
\bibfield{author}{\bibinfo{person}{Pritam Sarkar} {and} \bibinfo{person}{Ali Etemad}.} \bibinfo{year}{2020}\natexlab{}.
\newblock \showarticletitle{Self-supervised ECG representation learning for emotion recognition}.
\newblock \bibinfo{journal}{\emph{IEEE Transactions on Affective Computing}} \bibinfo{volume}{13}, \bibinfo{number}{3} (\bibinfo{year}{2020}), \bibinfo{pages}{1541--1554}.
\newblock


\bibitem[Siffer et~al\mbox{.}(2017)]%
        {siffer2017anomaly}
\bibfield{author}{\bibinfo{person}{Alban Siffer}, \bibinfo{person}{Pierre-Alain Fouque}, \bibinfo{person}{Alexandre Termier}, {and} \bibinfo{person}{Christine Largouet}.} \bibinfo{year}{2017}\natexlab{}.
\newblock \showarticletitle{Anomaly detection in streams with extreme value theory}. In \bibinfo{booktitle}{\emph{Proceedings of the 23rd ACM SIGKDD international conference on knowledge discovery and data mining}}. \bibinfo{pages}{1067--1075}.
\newblock


\bibitem[Tian et~al\mbox{.}(2020)]%
        {tian2020makes}
\bibfield{author}{\bibinfo{person}{Yonglong Tian}, \bibinfo{person}{Chen Sun}, \bibinfo{person}{Ben Poole}, \bibinfo{person}{Dilip Krishnan}, \bibinfo{person}{Cordelia Schmid}, {and} \bibinfo{person}{Phillip Isola}.} \bibinfo{year}{2020}\natexlab{}.
\newblock \showarticletitle{What makes for good views for contrastive learning?}
\newblock \bibinfo{journal}{\emph{Advances in neural information processing systems}}  \bibinfo{volume}{33} (\bibinfo{year}{2020}), \bibinfo{pages}{6827--6839}.
\newblock


\bibitem[Tonekaboni et~al\mbox{.}(2020)]%
        {tonekaboni2020unsupervised}
\bibfield{author}{\bibinfo{person}{Sana Tonekaboni}, \bibinfo{person}{Danny Eytan}, {and} \bibinfo{person}{Anna Goldenberg}.} \bibinfo{year}{2020}\natexlab{}.
\newblock \showarticletitle{Unsupervised Representation Learning for Time Series with Temporal Neighborhood Coding}. In \bibinfo{booktitle}{\emph{International Conference on Learning Representations}}.
\newblock


\bibitem[Wang et~al\mbox{.}(2023)]%
        {wang2023networked}
\bibfield{author}{\bibinfo{person}{Dingsu Wang}, \bibinfo{person}{Yuchen Yan}, \bibinfo{person}{Ruizhong Qiu}, \bibinfo{person}{Yada Zhu}, \bibinfo{person}{Kaiyu Guan}, \bibinfo{person}{Andrew Margenot}, {and} \bibinfo{person}{Hanghang Tong}.} \bibinfo{year}{2023}\natexlab{}.
\newblock \showarticletitle{Networked time series imputation via position-aware graph enhanced variational autoencoders}. In \bibinfo{booktitle}{\emph{Proceedings of the 29th ACM SIGKDD Conference on Knowledge Discovery and Data Mining}}. \bibinfo{pages}{2256--2268}.
\newblock


\bibitem[Wang and Isola(2020)]%
        {wang2020understanding}
\bibfield{author}{\bibinfo{person}{Tongzhou Wang} {and} \bibinfo{person}{Phillip Isola}.} \bibinfo{year}{2020}\natexlab{}.
\newblock \showarticletitle{Understanding contrastive representation learning through alignment and uniformity on the hypersphere}. In \bibinfo{booktitle}{\emph{International Conference on Machine Learning}}. PMLR, \bibinfo{pages}{9929--9939}.
\newblock


\bibitem[Wang et~al\mbox{.}(2019)]%
        {wang2019implicit}
\bibfield{author}{\bibinfo{person}{Yulin Wang}, \bibinfo{person}{Xuran Pan}, \bibinfo{person}{Shiji Song}, \bibinfo{person}{Hong Zhang}, \bibinfo{person}{Gao Huang}, {and} \bibinfo{person}{Cheng Wu}.} \bibinfo{year}{2019}\natexlab{}.
\newblock \showarticletitle{Implicit semantic data augmentation for deep networks}.
\newblock \bibinfo{journal}{\emph{Advances in Neural Information Processing Systems}}  \bibinfo{volume}{32} (\bibinfo{year}{2019}).
\newblock


\bibitem[Williams(1992)]%
        {williams1992simple}
\bibfield{author}{\bibinfo{person}{Ronald~J Williams}.} \bibinfo{year}{1992}\natexlab{}.
\newblock \showarticletitle{Simple statistical gradient-following algorithms for connectionist reinforcement learning}.
\newblock \bibinfo{journal}{\emph{Machine learning}}  \bibinfo{volume}{8} (\bibinfo{year}{1992}), \bibinfo{pages}{229--256}.
\newblock


\bibitem[Woo et~al\mbox{.}(2021)]%
        {woo2021cost}
\bibfield{author}{\bibinfo{person}{Gerald Woo}, \bibinfo{person}{Chenghao Liu}, \bibinfo{person}{Doyen Sahoo}, \bibinfo{person}{Akshat Kumar}, {and} \bibinfo{person}{Steven Hoi}.} \bibinfo{year}{2021}\natexlab{}.
\newblock \showarticletitle{CoST: Contrastive Learning of Disentangled Seasonal-Trend Representations for Time Series Forecasting}. In \bibinfo{booktitle}{\emph{International Conference on Learning Representations}}.
\newblock


\bibitem[Xu et~al\mbox{.}(2018)]%
        {xu2018unsupervised}
\bibfield{author}{\bibinfo{person}{Haowen Xu}, \bibinfo{person}{Wenxiao Chen}, \bibinfo{person}{Nengwen Zhao}, \bibinfo{person}{Zeyan Li}, \bibinfo{person}{Jiahao Bu}, \bibinfo{person}{Zhihan Li}, \bibinfo{person}{Ying Liu}, \bibinfo{person}{Youjian Zhao}, \bibinfo{person}{Dan Pei}, \bibinfo{person}{Yang Feng}, {et~al\mbox{.}}} \bibinfo{year}{2018}\natexlab{}.
\newblock \showarticletitle{Unsupervised anomaly detection via variational auto-encoder for seasonal kpis in web applications}. In \bibinfo{booktitle}{\emph{Proceedings of the 2018 world wide web conference}}. \bibinfo{pages}{187--196}.
\newblock


\bibitem[Yan et~al\mbox{.}(2023)]%
        {yan2024reconciling}
\bibfield{author}{\bibinfo{person}{Yuchen Yan}, \bibinfo{person}{Baoyu Jing}, \bibinfo{person}{Lihui Liu}, \bibinfo{person}{Ruijie Wang}, \bibinfo{person}{Jinning Li}, \bibinfo{person}{Tarek Abdelzaher}, {and} \bibinfo{person}{Hanghang Tong}.} \bibinfo{year}{2023}\natexlab{}.
\newblock \showarticletitle{Reconciling competing sampling strategies of network embedding}.
\newblock \bibinfo{journal}{\emph{Advances in Neural Information Processing Systems}}  \bibinfo{volume}{36} (\bibinfo{year}{2023}).
\newblock


\bibitem[Yin et~al\mbox{.}(2022)]%
        {yin2022autogcl}
\bibfield{author}{\bibinfo{person}{Yihang Yin}, \bibinfo{person}{Qingzhong Wang}, \bibinfo{person}{Siyu Huang}, \bibinfo{person}{Haoyi Xiong}, {and} \bibinfo{person}{Xiang Zhang}.} \bibinfo{year}{2022}\natexlab{}.
\newblock \showarticletitle{Autogcl: Automated graph contrastive learning via learnable view generators}. In \bibinfo{booktitle}{\emph{Proceedings of the AAAI conference on artificial intelligence}}, Vol.~\bibinfo{volume}{36}. \bibinfo{pages}{8892--8900}.
\newblock


\bibitem[You et~al\mbox{.}(2020)]%
        {you2020design}
\bibfield{author}{\bibinfo{person}{Jiaxuan You}, \bibinfo{person}{Zhitao Ying}, {and} \bibinfo{person}{Jure Leskovec}.} \bibinfo{year}{2020}\natexlab{}.
\newblock \showarticletitle{Design space for graph neural networks}.
\newblock \bibinfo{journal}{\emph{Advances in Neural Information Processing Systems}}  \bibinfo{volume}{33} (\bibinfo{year}{2020}), \bibinfo{pages}{17009--17021}.
\newblock


\bibitem[You et~al\mbox{.}(2021)]%
        {you2021graph}
\bibfield{author}{\bibinfo{person}{Yuning You}, \bibinfo{person}{Tianlong Chen}, \bibinfo{person}{Yang Shen}, {and} \bibinfo{person}{Zhangyang Wang}.} \bibinfo{year}{2021}\natexlab{}.
\newblock \showarticletitle{Graph contrastive learning automated}. In \bibinfo{booktitle}{\emph{International Conference on Machine Learning}}. PMLR, \bibinfo{pages}{12121--12132}.
\newblock


\bibitem[Yue et~al\mbox{.}(2022)]%
        {yue2022ts2vec}
\bibfield{author}{\bibinfo{person}{Zhihan Yue}, \bibinfo{person}{Yujing Wang}, \bibinfo{person}{Juanyong Duan}, \bibinfo{person}{Tianmeng Yang}, \bibinfo{person}{Congrui Huang}, \bibinfo{person}{Yunhai Tong}, {and} \bibinfo{person}{Bixiong Xu}.} \bibinfo{year}{2022}\natexlab{}.
\newblock \showarticletitle{Ts2vec: Towards universal representation of time series}. In \bibinfo{booktitle}{\emph{Proceedings of the AAAI Conference on Artificial Intelligence}}, Vol.~\bibinfo{volume}{36}. \bibinfo{pages}{8980--8987}.
\newblock


\bibitem[Zerveas et~al\mbox{.}(2021)]%
        {zerveas2021transformer}
\bibfield{author}{\bibinfo{person}{George Zerveas}, \bibinfo{person}{Srideepika Jayaraman}, \bibinfo{person}{Dhaval Patel}, \bibinfo{person}{Anuradha Bhamidipaty}, {and} \bibinfo{person}{Carsten Eickhoff}.} \bibinfo{year}{2021}\natexlab{}.
\newblock \showarticletitle{A transformer-based framework for multivariate time series representation learning}. In \bibinfo{booktitle}{\emph{Proceedings of the 27th ACM SIGKDD conference on knowledge discovery \& data mining}}. \bibinfo{pages}{2114--2124}.
\newblock


\bibitem[Zhang et~al\mbox{.}(2022)]%
        {zhang2022self}
\bibfield{author}{\bibinfo{person}{Xiang Zhang}, \bibinfo{person}{Ziyuan Zhao}, \bibinfo{person}{Theodoros Tsiligkaridis}, {and} \bibinfo{person}{Marinka Zitnik}.} \bibinfo{year}{2022}\natexlab{}.
\newblock \showarticletitle{Self-supervised contrastive pre-training for time series via time-frequency consistency}.
\newblock \bibinfo{journal}{\emph{Advances in Neural Information Processing Systems}}  \bibinfo{volume}{35} (\bibinfo{year}{2022}), \bibinfo{pages}{3988--4003}.
\newblock


\bibitem[Zhao et~al\mbox{.}(2021)]%
        {zhao2021automated}
\bibfield{author}{\bibinfo{person}{Xiangyu Zhao}, \bibinfo{person}{Haochen Liu}, \bibinfo{person}{Wenqi Fan}, \bibinfo{person}{Hui Liu}, \bibinfo{person}{Jiliang Tang}, {and} \bibinfo{person}{Chong Wang}.} \bibinfo{year}{2021}\natexlab{}.
\newblock \showarticletitle{Automated Loss Function Search in Recommendations}. In \bibinfo{booktitle}{\emph{27th ACM SIGKDD Conference on Knowledge Discovery and Data Mining: KDD 2021}}. \bibinfo{pages}{1--9}.
\newblock


\bibitem[Zheng et~al\mbox{.}(2024a)]%
        {zheng2024mulan}
\bibfield{author}{\bibinfo{person}{Lecheng Zheng}, \bibinfo{person}{Zhengzhang Chen}, \bibinfo{person}{Jingrui He}, {and} \bibinfo{person}{Haifeng Chen}.} \bibinfo{year}{2024}\natexlab{a}.
\newblock \showarticletitle{MULAN: Multi-modal Causal Structure Learning and Root Cause Analysis for Microservice Systems}. In \bibinfo{booktitle}{\emph{Proceedings of the ACM on Web Conference 2024}}. \bibinfo{pages}{4107--4116}.
\newblock


\bibitem[Zheng et~al\mbox{.}(2021)]%
        {zheng2021deeper}
\bibfield{author}{\bibinfo{person}{Lecheng Zheng}, \bibinfo{person}{Dongqi Fu}, \bibinfo{person}{Ross Maciejewski}, {and} \bibinfo{person}{Jingrui He}.} \bibinfo{year}{2021}\natexlab{}.
\newblock \showarticletitle{Deeper-GXX: deepening arbitrary GNNs}.
\newblock \bibinfo{journal}{\emph{arXiv preprint arXiv:2110.13798}} (\bibinfo{year}{2021}).
\newblock


\bibitem[Zheng et~al\mbox{.}(2024b)]%
        {zheng2024heterogeneous}
\bibfield{author}{\bibinfo{person}{Lecheng Zheng}, \bibinfo{person}{Baoyu Jing}, \bibinfo{person}{Zihao Li}, \bibinfo{person}{Hanghang Tong}, {and} \bibinfo{person}{Jingrui He}.} \bibinfo{year}{2024}\natexlab{b}.
\newblock \showarticletitle{Heterogeneous Contrastive Learning for Foundation Models and Beyond}.
\newblock \bibinfo{journal}{\emph{arXiv preprint arXiv:2404.00225}} (\bibinfo{year}{2024}).
\newblock


\bibitem[Zheng et~al\mbox{.}(2022)]%
        {zheng2022contrastive}
\bibfield{author}{\bibinfo{person}{Lecheng Zheng}, \bibinfo{person}{Jinjun Xiong}, \bibinfo{person}{Yada Zhu}, {and} \bibinfo{person}{Jingrui He}.} \bibinfo{year}{2022}\natexlab{}.
\newblock \showarticletitle{Contrastive learning with complex heterogeneity}. In \bibinfo{booktitle}{\emph{Proceedings of the 28th ACM SIGKDD Conference on Knowledge Discovery and Data Mining}}. \bibinfo{pages}{2594--2604}.
\newblock


\bibitem[Zheng et~al\mbox{.}(2023)]%
        {zheng2023fairness}
\bibfield{author}{\bibinfo{person}{Lecheng Zheng}, \bibinfo{person}{Yada Zhu}, {and} \bibinfo{person}{Jingrui He}.} \bibinfo{year}{2023}\natexlab{}.
\newblock \showarticletitle{Fairness-aware multi-view clustering}. In \bibinfo{booktitle}{\emph{Proceedings of the 2023 SIAM International Conference on Data Mining (SDM)}}. SIAM, \bibinfo{pages}{856--864}.
\newblock


\bibitem[Zhou et~al\mbox{.}(2020)]%
        {zhou2020data}
\bibfield{author}{\bibinfo{person}{Dawei Zhou}, \bibinfo{person}{Lecheng Zheng}, \bibinfo{person}{Jiawei Han}, {and} \bibinfo{person}{Jingrui He}.} \bibinfo{year}{2020}\natexlab{}.
\newblock \showarticletitle{A data-driven graph generative model for temporal interaction networks}. In \bibinfo{booktitle}{\emph{Proceedings of the 26th ACM SIGKDD International Conference on Knowledge Discovery \& Data Mining}}. \bibinfo{pages}{401--411}.
\newblock


\bibitem[Zhou et~al\mbox{.}(2021)]%
        {zhou2021informer}
\bibfield{author}{\bibinfo{person}{Haoyi Zhou}, \bibinfo{person}{Shanghang Zhang}, \bibinfo{person}{Jieqi Peng}, \bibinfo{person}{Shuai Zhang}, \bibinfo{person}{Jianxin Li}, \bibinfo{person}{Hui Xiong}, {and} \bibinfo{person}{Wancai Zhang}.} \bibinfo{year}{2021}\natexlab{}.
\newblock \showarticletitle{Informer: Beyond efficient transformer for long sequence time-series forecasting}. In \bibinfo{booktitle}{\emph{Proceedings of the AAAI conference on artificial intelligence}}, Vol.~\bibinfo{volume}{35}. \bibinfo{pages}{11106--11115}.
\newblock


\end{thebibliography}
}

\newpage
\appendix
\section{Data Augmentations}
Data augmentations transform the input data into different but related views, which are the cornerstones of a CLS.
In general, each augmentation is associated with a parameter $p\in[0,1]$.
Let $\mathbf{x}\in\mathbb{R}^{T\times c}$ be the input time series, where $T$ and $c$ are the length and the number of variables, then the details of data augmentations are given below.

\paragraph{Resizing (length)}
$p$ refers to the standard deviation of the Gaussian noise.
The final length is given by 

\paragraph{Rescaling (amplitude)}
$p$ refers to the standard deviation of the Gaussian noise.
The final amplitude is given by 

\paragraph{Jittering}
$p$ refers to the standard deviation of the Gaussian noise: $\mathbf{x}=\mathbf{x}+\mathcal{N}(0,p)$.

\paragraph{Point Masking}
$p$ refers to the ratio of the input data points to be masked: $\mathbf{x}=\mathbf{x}\odot\mathbf{m}$, where $\mathbf{m}[t]\sim Bernoulli(1-p)$.

\paragraph{Frequency Masking}
$p$ refers to the ratio of the frequencies to be masked.
We first apply Discrete Fourier Transform (DFT) over $\mathbf{x}$ to obtain its frequency representation $\mathbf{x}_f=DFT(\mathbf{x})$.
Next, we randomly mask out $p$ frequencies $\mathbf{x}_f=\mathbf{x}_f\odot\mathbf{m}$, where $\mathbf{m}[t]\sim Bernoulli(1-p)$.
Finally, we transform $\mathbf{x}_f$ back to the time domain via Inverse Discrete Fourier Transform (IDFT): $\mathbf{x}=IDFT(\mathbf{x}_f)$.

\paragraph{Random Cropping}
$p=\frac{T'}{T}$, where $T$ and $T'$ are the lengths of the input $\mathbf{x}$, and the length of the common segment shared by two cropped sub-sequences $\mathbf{x}_1$ and $\mathbf{x}_2$. 
Specifically, $\mathbf{x}$ is cropped into two sub-sequences with a shared segment, i.e., $\mathbf{x}_1=\mathbf{x}[t_1:t_1']$ and $\mathbf{x}_2=\mathbf{x}[t_2:t_2']$, where $t_1\leq t_2< t_1'\leq t_2'$ and $\mathbf{x}_c=\mathbf{x}[t_2:t_1']$ is shared by ${\mathbf{x}}_1$ and ${\mathbf{x}}_2$.
Then $T'=t_1'-t_2$ and $p=\frac{T'}{T}$.

\paragraph{Orders of Data Augmentations}
The order of applying data augmentations also influences the learned embeddings \cite{cubuk2019autoaugment}.
For instance, the outcome of applying {point masking} followed by {random cropping} differs from the sequence of first {random cropping} and then {point masking}.
However, the ordering of other data augmentations, such as {resizing} and {rescaling}, does not have a significant influence on the data. 
Therefore, we design 5 different orders of applying data augmentations.
The details are presented in Table \ref{tab:order}.

\begin{table}[]
    \centering
    \scriptsize
    \begin{tabular}{c|c|c|c|c|c}
    \toprule
    ID. & Order-1 & Order-2 & Order-3 & Order-4 & Order-5\\
    \midrule
    1 & Resize & Resize & Resize & Resize & Resize \\
    2 & Rescale & Rescale & Rescale & Rescale & Crop\\
    3 & Freq. Mask & Freq. Mask & Freq. Mask & Crop & Rescale\\
    4 & Jitter & Jitter & Crop & Freq. Mask & Freq. Mask\\
    5 & Point Mask & Crop & Jitter& Jitter & Jitter\\
    6 & Crop & Point Mask & Point Mask & Point Mask & Point Mask\\
    \bottomrule
    \end{tabular}
    \caption{Orders of data augmentations.}
    \label{tab:order}
\end{table}





\section{More Experimental Setups}\label{sec:exp_setup_cont}
\subsection{Evaluation Metrics}
Our evaluation follows \cite{yue2022ts2vec}.
For each dataset, we first pre-train an encoder based on a CLS, and then train a downstream model, which is evaluated on the test data.
For classification, the downstream model is the SVM classifier with RBF kernel, and the evaluation metrics are accuracy (ACC) and F1.
For forecasting, the linear regression model with $l_2$ norm is used, and the evaluation metrics are Mean Squared Error (MSE) and Mean Absolute Error (MAE).
For anomaly detection, we follow the protocol used by \cite{yue2022ts2vec,ren2019time}.
The evaluation metrics are F1, precision, and recall.

\subsection{Baselines}
We consider the following baselines:
InfoTS \cite{luo2023time} adaptively learns optimal augmentations.
TS2Vec \cite{yue2022ts2vec} performs contrastive learning in an hierarchical manner.
TS-TCC \cite{eldeletime} uses temporal and contextual contrasting to train encoders.
CoST \cite{woo2021cost} performs augmentations in both temporal and frequency domains.
TNC \cite{tonekaboni2020unsupervised} treats adjacent samples as positive pairs and non-adjacent samples as negative pairs.
CPC \cite{oord2018representation} introduces InfoNCE. 
Self-EEG \cite{sarkar2020self} uses various transformations to augment the input time series.
SR \cite{ren2019time} detects anomaly points based on the spectral residual.
DONUT \cite{xu2018unsupervised} detect anomalies based on Variational Auto-Encoder (VAE) \cite{doersch2016tutorial}.
SPOT and DSPOT \cite{siffer2017anomaly} use the extreme value theory \cite{beirlant2006statistics} to find anomalies.
Note that GGS (Section \ref{sec:ggs}) is the generally good strategy found by \autocss, which has strong performance on different datasets and tasks.

\subsection{Implementation Details}
For the controller network, its hidden dimension size is 320, and MLPs are linear layers with a softmax activation.
We use a 10 layer dilated CNN as the encoder similar to \cite{yue2022ts2vec}, and the sizes of hidden dimension and the output embeddings are 64 and 320.
The learning rate of the controller is 0.0001.
We fix the maximum number of the iterations in phase 1 as 500, and $\alpha$ as 10.
For classification, forecasting, and anomaly detection, 
the tolerance $\epsilon$ are  0.001, 0.0001, and 0.001, 
and the validation metrics are ACC, MSE (48 horizons), and F1.
In pre-training, we fix the maximum length of the input time series as 2,000.

\end{document}